\documentclass[10pt,twocolumn,letterpaper]{article}
\usepackage[accsupp]{axessibility}  

\usepackage[iccvfinal]{iccv}              

\usepackage{amsmath}%
\PassOptionsToPackage{table}{xcolor}

\usepackage{xcolor}
\usepackage{colortbl} 
\usepackage{booktabs}   
\usepackage{multirow}   
\usepackage{amssymb}    
\usepackage{bbding}     
\usepackage{xspace}    

\definecolor{iccvblue}{rgb}{0.21,0.49,0.74}
\definecolor{nicegreen}{rgb}{0.1, 0.6, 0.2}

\usepackage[pagebackref,breaklinks,colorlinks,allcolors=iccvblue]{hyperref}

\newcommand\blfootnote[1]{%
  \begingroup
  \renewcommand\thefootnote{}\footnote{#1}%
  \addtocounter{footnote}{-1}%
  \endgroup
}

\def\name{\textsc{GenFlowRL}\xspace}

\vspace{-0.5cm}
\def\paperID{15540} 
\def\confName{ICCV}
\def\confYear{2025}

\title{\name: Shaping Rewards with Generative Object-Centric Flow \\in Visual Reinforcement Learning}

\vspace{-0.5cm}
\author{
Kelin Yu$^{*}$ \quad
Sheng Zhang$^{*}$ \quad
Harshit Soora \quad
Furong Huang \quad
Heng Huang \quad
Pratap Tokekar \quad
Ruohan Gao \\
University of Maryland, College Park \\
{\tt\small \{kyu85, shengz, hsoora, furongh, heng, tokekar, rhgao\}@umd.edu}
}

\begin{document}
\twocolumn[{%
\renewcommand\twocolumn[1][]{#1}%
\maketitle
\begin{center}
    \centering
    \captionsetup{type=figure}
    \vspace{-0.3cm}
\includegraphics[width=0.97\textwidth]{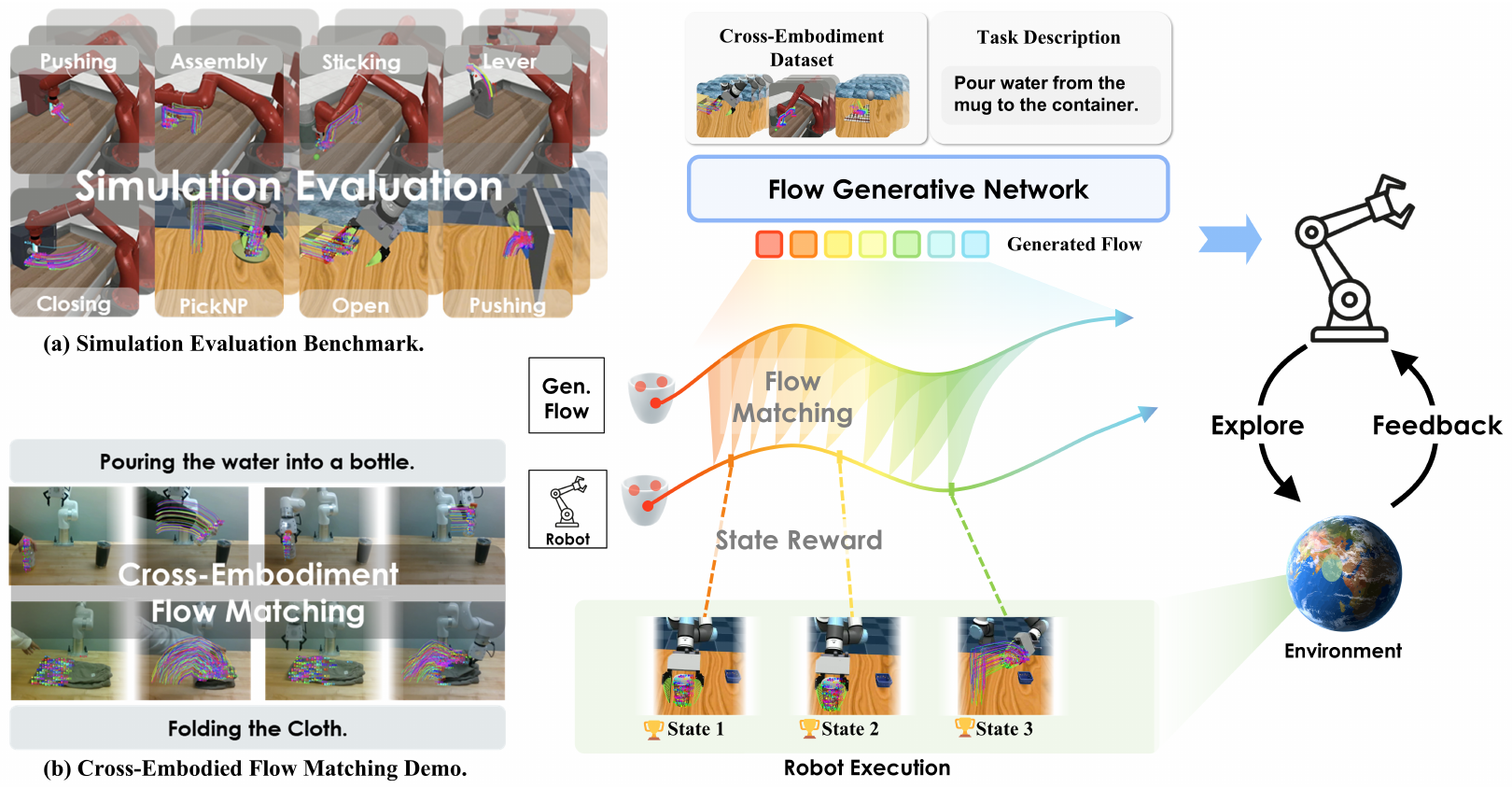}
    \vspace{-0.3cm}
    \caption{\textbf{Illustration of our \name framework}, which guides visuomotor RL policy by taking the generative object-centric flow as task motion prior (\textbf{Right}).
    In our proposed hybrid reward model, dense flow matching between online trajectories and flow prior, synergizing with sparse state-aware reward, facilitates efficient, robust, and generalizable policy learning.
    Our extensive evaluation includes 10 challenging simulation manipulation tasks (\textbf{Fig.~a}) and real-world cross-embodiment reward matchness probing experiments (\textbf{Fig.~b}).
    }
\label{fig:Diagram}
\vspace{-0.1cm}
\end{center}
}]
\begin{abstract}
\blfootnote{$^{*}$Equal contribution.}
Recent advances have shown that video generation models can enhance robot learning by deriving effective robot actions through inverse dynamics. However, these methods heavily depend on the quality of generated data and struggle with fine-grained manipulation due to the lack of environment feedback. While video-based reinforcement learning improves policy robustness, it remains constrained by the uncertainty of video generation and the challenges of collecting large-scale robot datasets for training diffusion models. To address these limitations, we propose \name, which derives shaped rewards from generated flow trained from diverse cross-embodiment datasets. This enables learning generalizable and robust policies from diverse demonstrations using low-dimensional, object-centric features. Experiments on 10 manipulation tasks, both in simulation and real-world cross-embodiment evaluations, demonstrate that \name effectively leverages manipulation features extracted from generated object-centric flow, consistently achieving superior performance across diverse and challenging scenarios.
Our Project Page: \href{link}{https://colinyu1.github.io/genflowrl}.
\end{abstract}    
\vspace{-0.3cm}
\section{Introduction}
\vspace{-0.15cm}
\label{sec:intro}

Recent advances in robot learning have demonstrated the potential of video generative foundation models in deriving future actions from generated robot frames using inverse dynamics~\citep{avdc, ardup, dreamite, this-that, flip}. This paradigm enables learning a universal policy across diverse scenarios with strong generalizability. 
However, existing methods solely rely on presumably perfect generated future frames for learning open-loop policies without actively interacting with the environment~\cite{dreamite, flip, ivideogpt}.
This leads to a lack of robustness, which is particularly critical in tasks that require fine-grained or contact-rich action prediction, where environmental feedback is indispensable. \par

Reinforcement Learning (RL), on the other hand, offers a complementary approach by allowing policies to interact with the environment and thus improve their robustness~\citep{yarats2022mastering, schulman2017proximalpolicyoptimizationalgorithms, haarnoja2018softactorcriticoffpolicymaximum}. 
Therefore, integrating video generation models with RL presents a promising direction~\citep{rank2reward, viper-24, Huang2023DiffusionReward, watching}, where the pre-trained video generative model serves as a conditional reward for RL training to enhance generalizability across diverse settings. 
Nevertheless, this approach is bottlenecked by the need for large-scale robotic data collection~\cite{robot-data-collection}, which remains prohibitively expensive and time-consuming.
Further, video generative models suffer from substantial artifacts, which degrade video quality and limit their effectiveness for reward shaping in RL training~\cite{vdm, flip}. \par 

To address these limitations while leveraging the complementary strengths of both video generative models and RL, we repurpose generated \textit{object-centric flow}~\cite{flow2act, flow-15, flip, hudor}, \ie, the 2D keypoints trajectory of objects, for reward shaping. 
Object-centric flow has facilitated robot learning by bridging the embodiment gap between robots and easy-to-collect human hand demonstrations~\cite{flow2act, hudor}. 
Compared to raw videos~\cite{viper, Huang2023DiffusionReward}, object-centric flow provides a much lower-dimensional and thus easier-to-generate representation that maintains essential manipulation-relevant features while abstracting away irrelevant details.
Additionally, expert object-centric flow enables direct reward shaping without the need for specialized representation learning~\cite{hudor}.
As shown in Table~\ref{tab:mani_rep}, object-centric flow offers fine-grained representations that better model manipulations of both deformable and articulated objects. These combined properties make it best-suited for \emph{generation} and \emph{reward shaping}. \par 

\begin{table*}[t]
    \centering
    \begin{minipage}{0.7\textwidth}
        \centering
        \begin{tabular}{@{}l|ccc|cc@{}}
        \toprule
        \multirow{2}{*}{\textbf{Representation}} 
        & \multicolumn{3}{c|}{\textbf{RL Compatibility}} 
        & \multicolumn{2}{c}{\textbf{Geo. Complexity}} \\
        \cmidrule{2-6}
        & {Low-Dim.} & {Cross-Emb.} & {Reward.} & {Deform.} & {Artic.} \\
        \midrule
        \rowcolor{red!5}
        Video Frames~\cite{viper, Huang2023DiffusionReward} &  &  &  & \checkmark & \checkmark \\
        \rowcolor{yellow!5}
        Gripper Keypoints~\cite{flowretrieval, point} & \checkmark &  & \checkmark & \checkmark & \checkmark \\
        \rowcolor{yellow!5}
        Active Region~\cite{ardup, MCR} &  &  &  & \checkmark & \checkmark \\
        \rowcolor{yellow!5}
        Trace~\cite{trace, sketch} & \checkmark &  &  & \checkmark & \checkmark \\
        \rowcolor{green!5}
        Object Pose~\cite{cimer, objdex} & \checkmark & \checkmark & \checkmark &  &  \\
        \rowcolor{green!5}
        Object Keypoints~\cite{rekep, viper} & \checkmark & \checkmark & \checkmark &  & \checkmark \\
        \rowcolor{green!5}
        Object Parts~\cite{part} &  & \checkmark &  &  & \checkmark \\
        \rowcolor{green!5}
        Object Flows~\cite{flow-15, flow2act, flip} & \checkmark & \checkmark & \checkmark & \checkmark & \checkmark \\
        \bottomrule
        \end{tabular}
        \label{tab:representations}
    \end{minipage}%
    \hfill
    \begin{minipage}{0.26\textwidth}
        \centering
        \includegraphics[width=\textwidth]{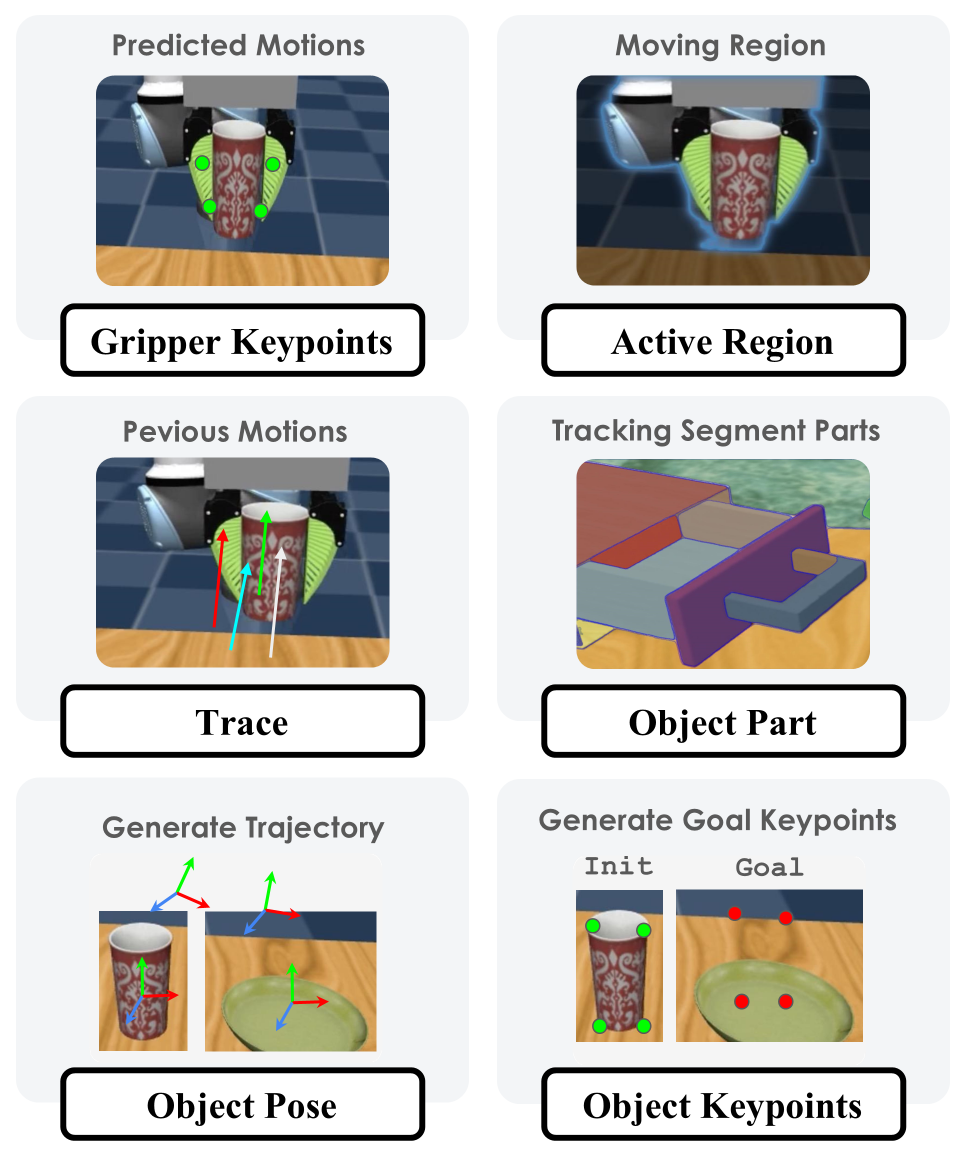}
    \end{minipage}
    \vspace{-0.3cm}
    \caption{
    \textbf{General comparisons among widely-adopted manipulation-centric representations for Robot Learning}, evaluated on their compatibility with RL and scalability to object geometric complexity. Three key dimensions are considered for \textit{RL compatibility}: (1) the dimensionality of the raw observational representation (Low-Dimensionality); (2) generalizability across different embodiments (Cross-Embodiment); and (3) supporting reward shaping from raw observations (Rewardability).
    Two key aspects are considered for \textit{geometric complexity}: (1) representability of deformable objects (Deformable); and (2) representability of articulated objects (Articulated).
    From the comparison, we observe that object-centric flow offers the highest flexibility and representational capacity and being well-suited for RL.}
    \label{tab:mani_rep}
    \vspace{-0.3cm}
\end{table*}

To this end, we propose \name, a novel framework that integrates object-centric flow generation with RL for robot manipulation.
Our approach harnesses strong and generalizable task priors from flow generative models while incorporating environmental feedback to overcome the limitations of open-loop policies. Specifically, we train a flow generative model on easy-to-collect cross-embodiment datasets to generate object-centric flows, which are then transformed into a condensed representation tailored for dense reward shaping.
Our hybrid reward model integrates the flow-derived dense rewards with sparse state-based rewards, leveraging both task and environment priors to learn specific flow-derived robot policy. For the flow-derived policy design, the generated flow serves as future motion conditions to improve policy robustness, while the initial 3D keypoints provide spatial context that enhances the reasoning over 2D flows. \par

Experiments on 10 manipulation tasks across two benchmarks demonstrate that our \name framework achieves faster convergence and superior performance, outperforming both flow-based imitation learning and video-guided RL methods in learning expert-like behaviors.
Our comparative analysis against manipulation-centric representations demonstrates the unique advantages of our adapted flow representation for fine-grained manipulation tasks.
Moreover, our real-world case study on a robot arm validates the effectiveness of our flow-derived reward model for cross-embodiment human-to-robot transfer. \par 

Our \textbf{key contributions} are three-fold:
\texttt{(1)} We propose \name, a novel framework that learns flow-derived RL policies from cross-embodiment videos for robot manipulation via our proposed flow-derived reward model, overcoming the limitations of the existing video-based RL paradigm.
\texttt{(2)} We comprehensively analyze various object-centric representations for practical insights, highlighting our effectiveness and efficiency. 
\texttt{(3)} Extensive evaluation on 10 challenging fine-grained manipulation tasks, both simulation and real-world cross-embodiment experiments, demonstrates the effectiveness of our method. \par

\vspace{-0.3cm}
\section{Related Work}
\label{sec:related}
\vspace{-0.15cm}

\noindent\textbf{Generative Models for Robot Learning.} 
Extensive work has explored the utility of various generative models for robot learning, \eg, LLMs~\cite{gpt4, llama}, VLMs~\cite{gpt4, clip}, Diffusion Models~\cite{animatediff}. 
These models have been widely adopted in planning~\cite{codeaspolicy, palme}, reward generation~\cite{text2reward, eureka, league}, and policy learning~\cite{chi2024diffusionpolicyvisuomotorpolicy}. 
Recent advancements in foundational video generative models, \eg, Stable Video Diffusion~\cite{stablediffusion}, have paved the way for leveraging generated future frames to learn universal robot policies~\cite{avdc, ardup, this-that, dreamite}. 
However, existing approaches are hindered by the low quality of generated videos and their reliance on open-loop policies. 
In contrast, our approach learns closed-loop policies through integration of robust object-centric flow and RL, incorporating active interaction with the environment. \par 

\vspace{0.05in}

\noindent\textbf{Reward Generation for Reinforcement Learning.} 
Prior efforts to address the prohibitive cost of manually designing task-specific rewards include inverse RL~\cite{airl, gail}, reward generation framework with large models~\cite{LIV, text2reward, eureka, league, rlvlm} for both planning and RL, and video-based RL~\cite{viper, Huang2023DiffusionReward, rank2reward, watching}.
Recent video-based RL approaches utilize video generative models for dense reward shaping~\citep{viper,rank2reward,Huang2023DiffusionReward}, achieving notable performance improvements.
However, these methods face challenges in capturing fine-grained manipulation-centric features directly from generated videos due to the inherent quality degradation and artifacts in video generation.
By contrast, our work bridges reinforcement learning and raw video with compressed low-dimensional object-centric flows for reward shaping, offering enhanced robustness and efficiency for fine-grained manipulation learning.

\vspace{0.05in}

\noindent\textbf{Manipulation-Centric Representations}\label{sec:discussion}
can be categorized into three types (Table~\ref{tab:mani_rep}): \colorbox{red!5}{\textit{raw representations}} (\eg, video frames~\cite{viper, Huang2023DiffusionReward}), \colorbox{yellow!5}{\textit{gripper-centric representations}} (\eg, gripper keypoints~\citep{point}, traces~\cite{trace}, active regions~\cite{ardup, MCR}), and \colorbox{green!5}{\textit{object-centric representations}} (\eg, object keypoints~\cite{rekep, iker}, 6D object poses~\cite{pose, objdex, cimer}, object parts~\cite{part}). 
Despite promising progress, gripper-centric representations rely on real robot demonstrations, limiting their ability to utilize easy-to-collect human hand demonstrations~\cite{flow2act, egomimic, mimictouch}, while most object-centric representations struggle with deformable and articulated objects due to insufficient granularity. \par

Although prior works investigate the potential of generated flows in imitation learning or trajectory optimization~\cite{flow2act, flow-15, flip, hudor}, our work differs in two primary aspects: \texttt{(1)} We introduce a condensed flow representation, $\delta$-flow, which encodes richer object dynamics to better handle fine-grained manipulations, as demonstrated in our comparative analysis; \texttt{(2)} 
In contrast to HuDor~\cite{hudor}, which employs a single expert flow for reward shaping in open-loop policy fine-tuning, we leverage generated flow as a cross-embodiment motion prior for dense reward shaping, enabling scalable closed-loop policy learning and achieving superior generalization and robustness across challenging manipulation tasks. \par 
\section{Methodology}
\label{sec:method}
To enable RL to learn from expert motion priors with better robustness and generalizability, we present \name, 
a framework that bridges the gap between generative foundation models and RL to tackle diverse manipulation tasks that demand fine-grained and contact-rich interactions. 
The key innovation of \name is our proposed object-centric flow-derived reward model for specific flow-derived policy learning. This reward model incentivizes exploration upon generated $\delta$-flow representations while mimicking expert motion priors from large-scale, cross-embodiment datasets. 
Additionally, it incorporates real-time feedback to enhance robustness against noise and potential inaccuracies in these priors. \par 

The overall architecture of our method is illustrated in Fig.~\ref{fig:Diagram}. Our framework comprises three main stages: \texttt{(1)} Task-conditioned object-centric flow generation, which produces object-centric flows conditioned on task descriptions (Sec.~\ref{sec:generation}).
\texttt{(2)} A hybrid reward model guides policy learning through continuous feedback from dense matching with our $\delta$-flow priors in addition to discrete sparse rewards from environmental and object states (Sec.~\ref{sec:rm}).
\texttt{(3)} Flow-conditioned policy learning, where we train a generalizable policy using our hybrid reward model, conditioned on $\delta$-flow priors (Sec.~\ref{sec:algo}). \par 

\begin{figure*}[t]
\vspace{-0.2cm}
    \centering
    \includegraphics[width=\linewidth]{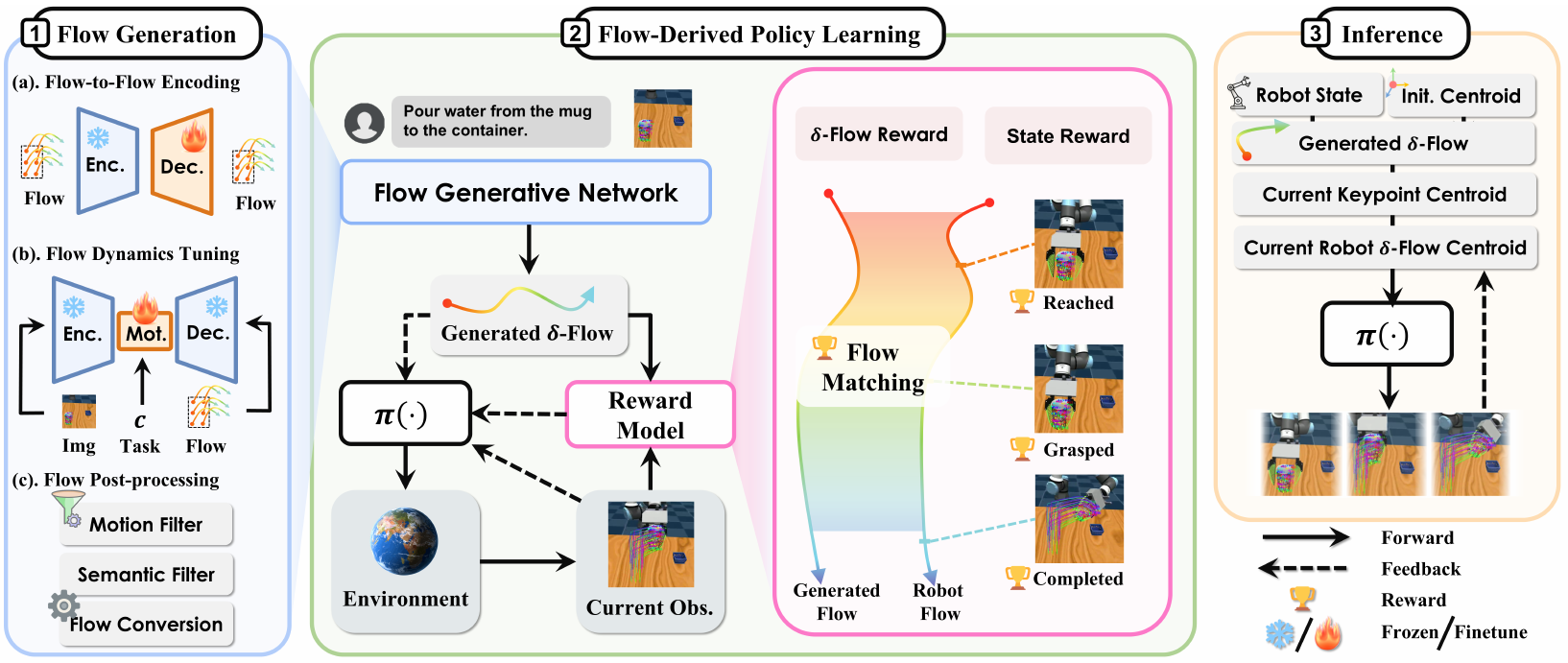}
    \caption{\textbf{Architectural overview of our proposed \name framework}, which encompasses the flow generation process (\textbf{left}), flow-derived policy learning (\textbf{middle}), and inference stage (\textbf{right}).
    In the \textit{object-centric flow generation} process (Sec.~\ref{sec:generation}), we: \texttt{(a)} adapt a pre-trained generative model decoder via flow-to-flow reconstruction; \texttt{(b)} fine-tune the latent motion module on flow generation conditioned on task descriptions and the initial keypoints; and
    \texttt{(c)} refine the generated flow with motion/semantic filtering and then convert it into our effective $\delta$-flow representation (Sec.~\ref{sec:rm}).
    In our \textit{flow-derived policy learning} stage, our hybrid reward model (Sec.~\ref{sec:rm}) guides policy learning through a dense $\delta$-flow matching reward derived from generated flows and a sparse state-aware reward derived from environmental interactions.
    During \textit{inference}, the policy executes 6D robot actions with conditions: robot state, initial 3D keypoint centroid, future steps of the generated $\delta$-flow, and current observation feedback.
    }
\vspace{-0.4cm}
\label{fig:Diagram}
\end{figure*}

\subsection{Object-Centric Flow Generation}
\label{sec:generation}
Adapting pre-trained video generative models to predict scene dynamics in the pixel space often leads to local artifacts, distortions, and factual inaccuracies that destabilize reinforcement learning for robotics tasks~\cite{dreamite, flip, ivideogpt}. 
In contrast, object-centric representation avoids these pitfalls thanks to its RL compatibility (see Tab.~\ref{tab:mani_rep}). 
\textit{Object-centric flow} represents the temporal motion of keypoints on a target object relative to the initial frame, abstracting away visual appearances~\cite{flow2act}.
Below, we describe how we generate object-centric flows by adapting video generative models~\cite{flow-15, flow2act, flip}. 
Notably, our empirical results indicate that while effective, this object-centric representation is suboptimal for RL, which motivates us to transform it into our $\delta$-flow representation (detailed in Sec.~\ref{sec:rm}). \par 

\noindent\textbf{Flow Generation Process}: 
Overall, the entire flow generation process encompasses three steps:
\texttt{(1)} \emph{flow dataset construction}: We convert an offline trajectory video dataset into an object-centric flow dataset by detecting the bounding box of the interactive object at the initial frame using Grounding-Dino~\cite{grounding-dino} and tracking temporal movements of uniformly sampled object keypoints with an off-the-shelf tracker~\cite{cotracker}. 
\texttt{(2)} \emph{generative model adaptation}: We adapt the pre-trained diffusion-based video generative model, AnimateDiff~\cite{animatediff}, into our flow generative model, conditioned on task descriptions in two fine-tuning stages. First, we fix its encoder and only finetune the decoder to adapt it to the flow dataset. Second, we only efficiently finetune the LoRA~\cite{lora} injected into the latent motion module to model the temporal dynamics of object flows. 
\texttt{(3)} \emph{post-processing of generated flows}: We apply motion and semantic filtering to reduce noise and ensure the sampled and generated keypoints reside within the object boundaries. \par 

\subsection{Flow-Derived Reward Model}
\label{rl_train}
\label{sec:rm}
The keypoints of the generated flow in the previous step offer dense temporal guidance from object-centric spatial features but still contain noisy and misleading trajectory cues.
Motivated by the above, we \textit{first} design a more robust representation, namely $\delta$-flow, which proves to be efficient and tailored to RL training.
\textit{Secondly}, we develop a hybrid reward model derived from flow, which combines a flow-matching reward---aligning current observations with the generated flow prior---with a sparse task-based reward. This reward model actively balances knowledge transfer and exploration by leveraging 
expert motion priors from the flow while following physically grounded on the sparse reward. 
As a result, our reward model enables the robot to robustly generalize to diverse manipulation settings and objects. \par

\noindent\textbf{$\delta$-Flow Construction}:
To further reduce the noises in generated object-centric flow, we propose to transform it into the condensed $\delta$-flow representation. It characterizes each timestep information of the original keypoint flow with three primary statistical estimates: the 2D centroid positions of object keypoints $\mathbf{\bar{P}}^t$, between-frame average translation of keypoints $\boldsymbol{\delta}^t_{tr}$, and between-frame average rotary transformation of keypoints $\boldsymbol{\delta}^t_{rot}$ at $t$-th step.
The definitions of these estimates are formulated as:
\begin{align}
    \mathbf{\bar{P}^t} &= \frac{1}{N} \sum_{i=1}^{N} \mathbf{P_i^t}, \quad
    \boldsymbol{\delta}_{\mathrm{tr}}^t = \mathbf{\bar{P}^t} - \mathbf{\bar{P}^1}, \\
    \boldsymbol{\delta}_{\mathrm{rot}}^t &= \frac{1}{N} \sum_{i=1}^N 
    \left[ \left( \mathbf{P_i^t} - \mathbf{\bar{P}^t} \right) \times \left( \mathbf{P_i^1} - \mathbf{\bar{P}^1} \right) \right],
\end{align}
where $\mathbf{P_i^t}$ denotes the 2D image coordinates of $i$-th keypoint on the $t$-th frame, and both $\boldsymbol{\delta}_{\mathrm{tr}}^t, \boldsymbol{\delta}_{\mathrm{rot}}^t$ are computed with $\mathbf{\bar{P}^1}$, the centroid of the $1^{st}$ frame.
For simplicity, we summarize relative object motions at $t$-th step as $\mathcal{T}^t = (\boldsymbol{\delta}_{\mathrm{tr}}^t; \boldsymbol{\delta}_{\mathrm{rot}}^t)$, which we refer to as \textbf{$\delta$-flow}.
Finally, we differentiate between two distinct $\delta$-flows: the robot execution or observational $\delta$-flow, denoted as $\mathcal{T}^t_R$, and the generated $\delta$-flow, $\mathcal{T}^t_G$. \par 

The $\delta$-flow representation provides a compact yet informative characterization of object 2D motion/dynamics.
With its Monte Carlo nature, it effectively reduces the negative impact of unreliable noisy keypoints predicted by the flow generation module.
This enhances the robustness and reliability of reward computation, ultimately improving the efficiency of RL training (see Sec.~\ref{exp:video}). \par 

\noindent\textbf{$\delta$-Flow Matching as Dense Reward}:
The generated $\delta$-flow, despite its simplicity, can readily serve as an effective continuous prior for object motion, guiding policy learning for coarse movements, which helps to overcome the sparse reward issue.
Thus, we aim to design a dense reward for matching executed trajectories with the $\delta$-flow, \ie, by quantifying the alignment between the observation flow induced by the online policy and the generated flow prior.
\textit{Specifically}, considering trajectory noise, we model the $t$-th timestep of generated and robot flows, $\mathcal{T}_G^t$ and $\mathcal{T}_R^t$, as samples drawn from their respective underlying probabilistic distributions, $P_{\mathcal{T}_R}^t$ and $P_{\mathcal{T}_G}^t$, which captures the underlying uncertainty in respective stochastic $\delta$-flow trajectories.
\textit{Furthermore}, we quantify the alignment between these flows with a distributional distance measure $\mathcal{D}$ that maps two distributions to a distance value.
Intuitively, the $t$-th step of the $\delta$-flow matching reward, $R_\delta^t$, should be inversely proportional to the distributional distance:
\begin{equation}
    R_\delta^t \propto -\mathcal{D}\Big(P_{\mathcal{T}_R^t}(\cdot| \mathbf{x}_{1:t}, \mathbf{P^1}, c), P_{\mathcal{T}_G^t}(\cdot| \mathbf{x}_{1}, \mathbf{P^1}, \phi, c)\Big)
    \label{eq:kld}
\end{equation}
where $\mathbf{x}_{1:t}$ denotes the observations up to timestep $t$. Both robot and generated flows are conditioned on the task $c$ and keypoints of the $1^{st}$ frame. The generated flow comes from the flow generative model parameterized by $\phi$. \par 

Our objective is to align two flows by maximizing the per-timestep reward, or equivalently, minimizing the distance between the two $\delta$-flow distributions. 
In practice, we adopt a simplified yet general assumption, \ie, Gaussian distribution with tied variance for two flows and Kullback–Leibler divergence (KLD) as our distance measure $\mathcal{D}$.
Under this assumption\footnote{The KLD between two Gaussians with tied variance is weighted $L_2$-norm distance between their means.}, the alignment between two $\delta$-flows reduces to matching their respective means, where the robot flow mean $\mathcal{T}_R$ serves as the target for the generated flow mean $\mathcal{T}_G$.
Thus, the normalized flow-matching reward at the $t$-th step, instantiated from eq.~(\ref{eq:kld}), is formulated as:
\begin{align}
R_{\delta}^t &= 1 - \text{clip} \left( \frac{\left(\mathcal{T}_R^t - \mathcal{T}_G^{t}\right)^2}{C}, 0, 1 \right)
\end{align}
where $C$ is a scaling parameter to control the variance of the approximation. 
To improve RL training stability, we adopt the clipping trick~\cite{schulman2017proximalpolicyoptimizationalgorithms} and normalize this reward to $[0, 1]$. \par 

\noindent\textbf{Overall Reward Model}:
In addition to tracking the expert motion prior in the generated $\delta$-flow, we propose an object \textit{state-aware reward} derived from the environmental feedback to reshape the sparse reward that is only grounded on task completion. 
This state-aware reward not only accelerates RL training but also provides the policy task-specific information, increasing its capability to accomplish the tasks. 
By incorporating $\delta$-flow and state-aware rewards, our entire reward model is designed as:
\begin{align}
R^t =
\begin{cases}
\alpha \cdot \left( 1 - \tanh(\tau \cdot d_{\text{grip}}) \right), & \text{if}\; d_{\text{grip}}>0, \\
\alpha, & \text{if finish subgoal}, \\
\alpha + \beta \cdot R_{\delta}^t, & \text{After subgoal}, \\
1.0, & \text{if completed}.
\end{cases}
\label{eq:reward}
\end{align}
where $d_{grip}$ is the distance between the gripper to the object. In practice, we set $\alpha=0.25, \beta=0.75$ and the temperature $\tau=10$. 
This task-agnostic design of our reward signal ensures its broad generalizability across diverse tasks. \par 

\subsection{Policy Design}
\label{sec:algo}
In this section, we detail our policy design grounded on the flow-derived reward model.
Different from existing RL policies~\cite{hudor, viper, Huang2023DiffusionReward}, our policy accepts a low-dimensional generated $\delta$-flow that encodes planned future keypoints as reference object motion priors. 
These priors serve as conditional inputs for next-action generation~\cite{MBA}. 
Additionally, the policy incorporates 3D object priors with the initial 3D reference keypoints in the initial frame to enhance its spatial understanding of how 2D flow transformations correspond to 3D object position changes. \par 

Specifically, we define the input space of our policy at $t$-th step as: \texttt{(1)} the current robot state $\mathbf{s}_t$, \texttt{(2)} the current centroid positions of observational object keypoints $\mathbf{\bar{P}}^t$, \texttt{(3)} the current observational object $\delta$-flow during robot execution $\mathcal{T}_R^t$, \texttt{(4)} the $k$-step lookahead generated keypoint centroid $\mathbf{\bar{P}}_G^{t+1: t+k}$, \texttt{(5)} the $k$-step lookahead generated $\delta$-flow $\mathcal{T}_G^{t+1: t+k}$, and \texttt{(6)} the 3D centroid positions at the $1^{st}$ frame $\mathbf{\bar{P}}_{\text{3d}}^1$. Formally, our policy is formulated as:
\begin{align}
\mathbf{a}_t &= \pi\Big(\mathbf{s}_t,\, \mathbf{\bar{P}}^t,\, \mathcal{T}_R^t,\, \mathbf{\bar{P}}_G^{t+1:t+k},\, \mathcal{T}_G^{t+1:t+k},\, \mathbf{\bar{P}}_{\text{3d}}^1\Big)
\end{align}
where the output action $\mathbf{a}_t$ is a 6D pose displacement used for RL exploration and policy learning, which is then transformed into joint robot commands by theinverse kinematics (IK). 
During policy training, we optimize the policy by maximizing our flow-derived reward (e.q.~\ref{eq:reward}), leveraging the DrQv2~\citep{yarats2022mastering} algorithm and its experience replay strategy. \par

\section{Experiments}
\vspace{-0.1cm}
\label{sec:experiment}
We are motivated to answer the following key questions: 
\textbf{(Q1)} Compared to flow-based Imitation Learning (IL), how effective and robust is our method, \ie, RL with our flow-derived reward model (Sec.~\ref{sec:rl_with_flows})? 
\textbf{(Q2)} How effective is $\delta$-flow against other manipulation-centric representations for RL reward shaping (Sec.~\ref{exp:rep})?
\textbf{(Q3)} What are the advantages of $\delta$-flow over other object-centric representations in the context of RL (Sec.~\ref{exp:rep})? 
\textit{Furthermore}, we also conduct ablation studies on policy input (Sec.~\ref{sec:ablation}) and present a real-world case study on reward matchness between human hand demonstrations and robot execution (Sec.~\ref{sec:real_world}).
Our evaluation and analyses are based on 10 challenging robotic manipulation tasks from two benchmarks. \par 

\begin{figure}[t]
\vspace{-0.2cm}
    \centering
    \includegraphics[width=0.5\textwidth]{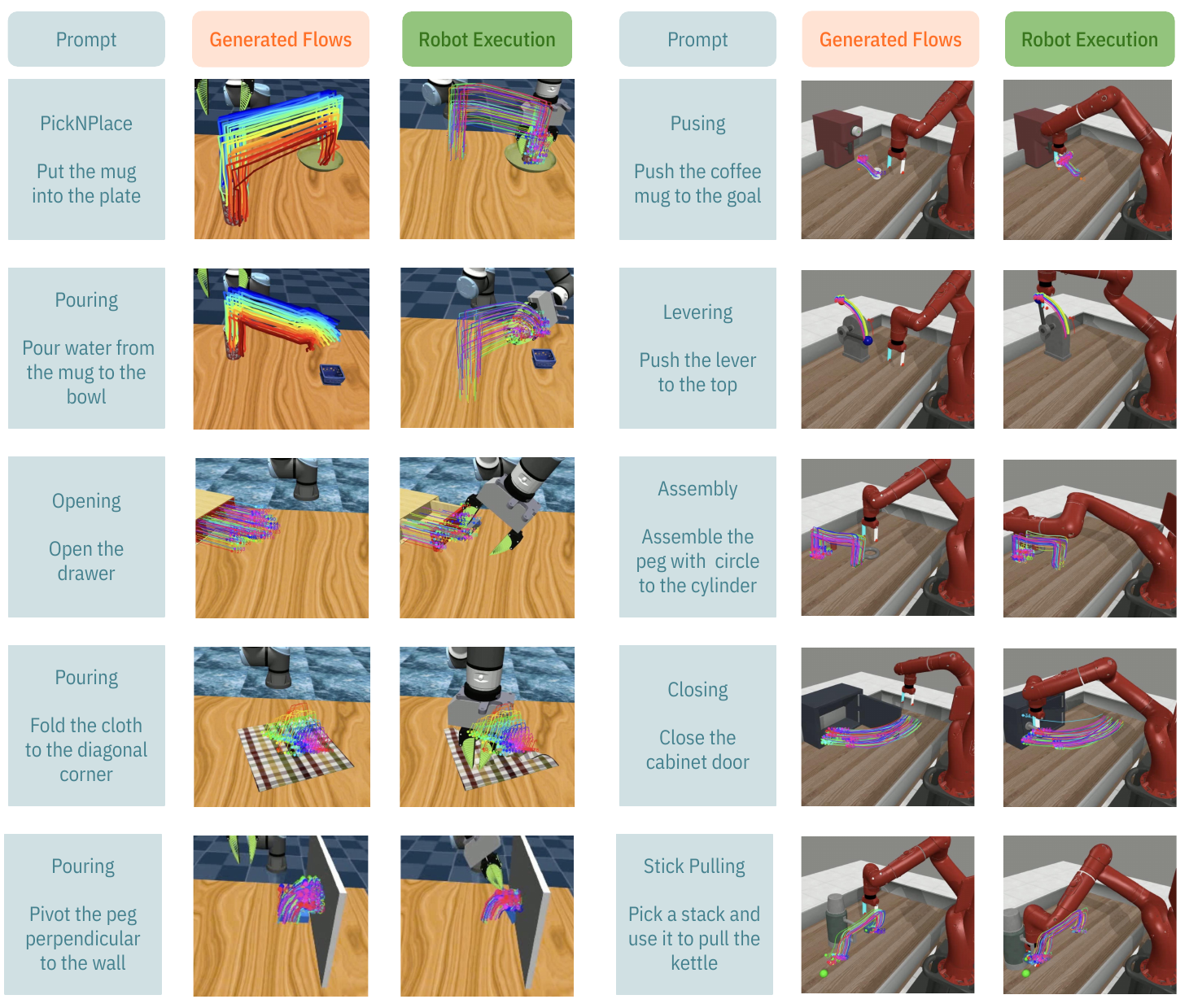}
    \vspace{-0.3cm}
    \caption{\textbf{Overview} of all task settings and robot execution demonstrations. \textbf{Left}: tasks in Im2Flow2Act~\cite{flow2act} benchmark. \textbf{Right}: tasks from MetaWorld~\cite{metaworld} benchmark. Columns denote textual task prompts, generated flows, and robot execution flows.}
\label{fig:task}
\vspace{-0.5cm}
\end{figure}

\vspace{0.05cm}

\noindent\textbf{Implementation Details:} 
We fine-tune the pre-trained StableDiffusion-1.5~\cite{stablediffusion} with two stages into our flow generative model. 
The cross-embodiment fine-tuning dataset contains 12K trajectories from 10 different tasks across three different embodiments: sphere robot, UR5, and Sawyar. See our Supp. for more dataset and training details. \par 
\vspace{0.05cm}

\noindent\textbf{Tasks Settings:} We construct our benchmark from MetaWorld~\cite{metaworld}, Im2Flow2Act~\cite{flow2act}, and our implementation.
We sample four tasks—\textit{PickNPlace}, \textit{Pouring}, \textit{Opening}, and \textit{Folding} from Im2Flow2Act. Besides, we implement a novel contact-rich task, \textit{Pivoting}, in this environment, which aims to 
reorient an object from a tabletop onto a vertical wall with pivoting motions.
These tasks require contact-rich and deformable object manipulation and thus are tailored to ablation with flow-based IL (in Sec.~\ref{sec:rl_with_flows}). 
For MetaWorld, we sample five of the most challenging tasks—\textit{Assembly}, \textit{Close Door}, \textit{Coffee Push}, \textit{Lever Pull}, and \textit{Stick Pull}.
All these tasks are designed based on Mujoco Engine~\cite{mujoco}. 
An UR5e robot and an Sawyar robot are used to collect data and do online exploration during RL training, respectively. 
Tasks are detailed in Fig.~\ref{fig:task} and Supp. \par 

\vspace{-0.05cm}
\subsection{How Effective and Robust is RL with $\delta$-Flow?}
\label{sec:rl_with_flows}
\vspace{-0.05cm}
To investigate the potential of flow-based RL, we evaluate our \name against two strong flow-based IL methods on five established robotic manipulation tasks in two setups. 
The comparison results are presented in Table~\ref{tab:method_results}. \par

\begin{table*}[h]
\centering
\begin{tabular}{l|c c c c c|c c c c c}
\toprule
 & \multicolumn{5}{c|}{\textbf{Demonstration-Conditioned}} & \multicolumn{5}{c}{\textbf{Language-Conditioned}} \\
 & PickNP. & Pour & Open & Fold & Pivot & PickNP. & Pour & Fold & Fold & Pivot \\ \midrule
\textbf{Heuristic}~\cite{flow-15} & 70 & 50 & 30 & 0 & 0 & / & / & / & / & / \\
\textbf{Im2Flow2Act}~\cite{flow2act} & 100 & 95 & 95 & 90 & 60 & 90 & 85 & 90 & 35 & 45 \\
\textbf{\name (Ours)} & \textbf{100} & \textbf{100} & \textbf{100} & \textbf{95} & \textbf{90} & \textbf{95} & \textbf{95} & \textbf{95} & \textbf{80} & \textbf{85} \\ \bottomrule
\end{tabular}
\vspace{-0.05in}
\caption{\textbf{Performance comparison for reward model ablations} across various methods on 5 challenging simulation tasks under two setups: demonstration-conditioned and language-conditioned execution. Scores reported in the success rate.}
\label{tab:method_results}
\vspace{-0.4cm}
\end{table*}

\vspace{0.05cm}

\noindent \textbf{Baselines:} We compare our framework with two baselines: \texttt{(1)} \textit{Heuristic Policy}: A heuristic action policy that selects object contact points in the $1^{st}$ frame and applies pose estimation to the object 3D flow in the following steps. This baseline is adapted from General Flow~\cite{flow-15}, and we provide the ground-truth 3D flow as input. 
\texttt{(2)} \textit{Im2Flow2Act}: A two-stage flow-based IL method~\cite{flow2act},  which also generates 2D flows but, instead of online RL training, it trains a task-agnostic flow-based diffusion policy using simulated heuristic actions.

\vspace{0.05cm}

\noindent \textbf{Evaluation}: 
Following existing practice~\cite{flow2act}, we evaluate in two setups: \textit{Demo-conditioned execution,} 
which evaluates each flow-based policy conditioned on the oracle flow extracted from expert demos in the dataset
and \textit{Language-conditioned execution}, 
where the flows guiding the policy are generated by conditions on the first frame and task language descriptions. \par 

\vspace{0.05cm}

\noindent \textbf{Key Findings:} 
The motion complexity of deformable objects and inevitable linguistic ambiguity introduce noises and defects in flow generation, which leads to notable performance degeneration in IL-based methods. 
Our method effectively overcomes this challenge through a novel policy design that ensures training-inference consistency and noise tolerance. \par 

As demonstrated in Table~\ref{tab:method_results}, our method clearly outperforms all baselines across all five tasks, with particularly strong performance on the challenging \textit{Folding} and \textit{Pivoting} tasks under language conditions. This superiority in robustness stems from our condensed $\delta$-flow representation in both policy design and reward model, which effectively filters out noise while retaining sufficient spatiotemporal task information. 
This noise-tolerant representation serves as reliable input to our policy. 
As opposed to baselines, it enables us to maintain training-inference consistency by employing generated flow as policy input for both training and inference rather than relying on costly expert flow. 
Combined with the inherent benefit of environmental feedback in RL, this design ensures robust performance while aligning with real-world requirements and scenarios. \par 

In contact-rich manipulation tasks, \eg, \textit{Pivoting}, the robot must track complex expert trajectories involving diverse contacts in unstructured environments. 
By actively interacting with the environment, our policy collects a broader distribution of trajectories, achieving closer alignment with the expert motion prior. \par
\vspace{-0.1cm}
\subsection{How Effective is $\delta$-Flow for RL?}
\label{exp:video}
\begin{figure*}[t]
    \centering
    \includegraphics[width=1\textwidth]{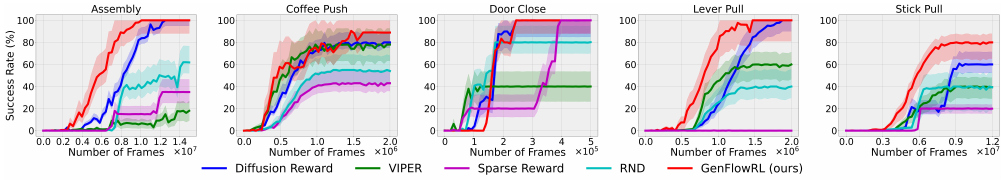}
    \vspace{-0.25in}
    \caption{Comparing with other video-based reward models. The shaded area represents the standard deviation for three random seeds.}
\label{fig:result_dr}
\vspace{-0.3cm}
\end{figure*}

\vspace{-0.05cm}
To analyze the advantages of our $\delta$-flow in RL, we compare our method with four state-of-the-art RL-based baselines with different representations on five robotic manipulation tasks in MetaWorld~\cite{metaworld}. Results are displayed in Fig.~\ref{fig:result_dr}. \par

\vspace{0.05cm}
\noindent \textbf{Baselines include: }
\texttt{(1)} \textit{Pure Sparse Reward (PSR)} purely adopts a state-aware sparse reward, which aims to demonstrate the effectiveness of dense reward shaping methods. 
\texttt{(2)} \textit{VIPER}~\cite{viper} 
introduces the next-frame log-likelihood of agent observation as RL reward from an adapted video generative model from VideoGPT~\cite{videogpt}.
\texttt{(3)} \textit{Diffusion Reward}~\cite{Huang2023DiffusionReward} is the state-of-the-art video-based reward shaping method with a conditional entropy reward upon VIPER~\cite{viper} for expert-like trajectories.
\texttt{(4)} \textit{Random Network Distillation (RND)}~\cite{rnd} encourages exploration with a novelty-seeking reward, different from reward shaping with pre-trained generative models. \par 

\vspace{0.05cm}

\noindent \textbf{Key Findings: }
Compared with non-pretrained methods like PSR and RND, reward shaping with a model pre-trained from expert demonstrations achieves superior performance. While these baselines perform competitively well on simple tasks, \eg, \textit{Door Close} and \textit{Coffee Push}, they struggle in more challenging manipulation tasks such as \textit{Assembly}. The primary limitation of these approaches is the absence of expert-guided rewards, which are crucial for incentivizing the exploration of expert-like behaviors. \par 

Compared with pre-trained video generation models, the generated object-centric flow shows better training efficacy, indicating that extracting essential manipulation-relevant features from complex video distributions is particularly challenging for intricate tasks. 
In contrast, our framework leverages $\delta$-flow as a more robust representation with manipulation-centric spatiotemporal features, effectively providing expert motion priors during exploration. \par 

\vspace{-0.05cm}
\subsection{$\delta$-Flow Achieves The Best Performance}
\label{exp:rep}
\begin{figure}[t]
    \centering
    \includegraphics[width=0.48\textwidth]{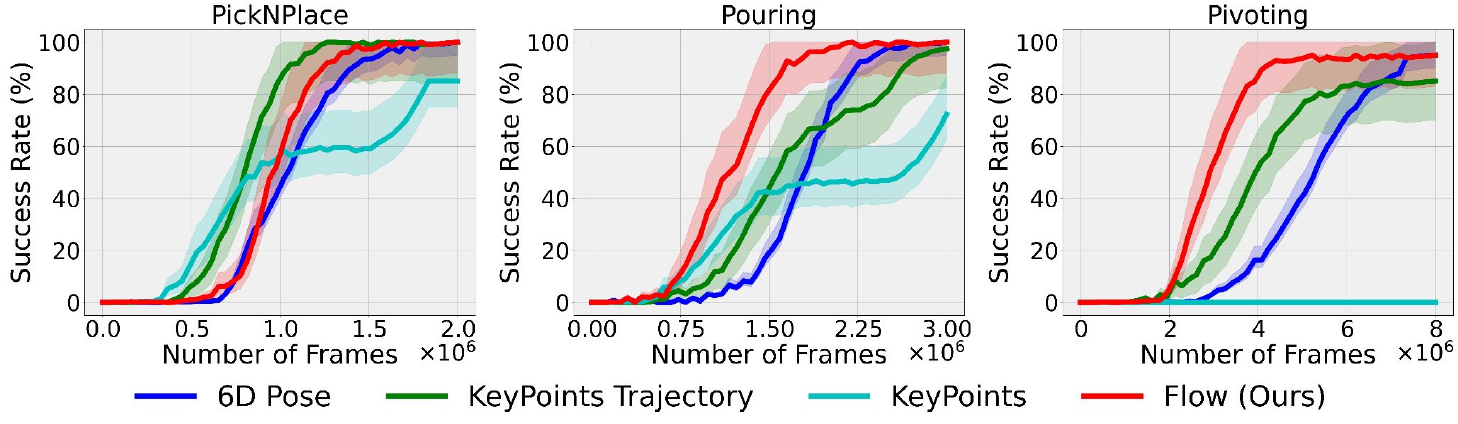}
    \vspace{-0.2cm}
    \caption{\textbf{Representation Evaluations.} The comparison of RL training results with different object-centric representations in three tasks. The shaded area represents the standard deviation for three random seeds.}
\label{fig:result_rep}
\vspace{-0.1cm}
\end{figure}

In this section, we aim to identify the most effective object-centric representation for reward shaping. We evaluate the training success rate of our framework using three alternative object-centric representations across three tasks in Im2Flow2Act. Figure~\ref{fig:result_rep} shows a comparison between our framework and the baselines.

\noindent \textbf{Baselines:} We compare the object-centric flow with three other expressive representations: \texttt{(1)} \textit{6D object poses~\cite{pose, cimer, objdex}}, which uses the object's 6D pose trajectory for reward shaping. \texttt{(2)} \textit{3D keypoints~\cite{rekep, iker}}, which manually tracks several 3D keypoints on the object for reward shaping. 
\texttt{(3)} \textit{3D keypoints trajectory}, which extends 3D keypoint-based reward shaping by incorporating keypoint trajectories in addition to the initial and goal positions. We exclude other manipulation-centric representations from Table~\ref{tab:mani_rep} due to their unsuitability for reward computation or incompatibility with cross-embodiment datasets. \par 

\noindent \textbf{Evaluation:} To evaluate different object-centric representations, we use ground-truth demonstrations as policy input. We select \textit{PickNPlace}, \textit{Pouring}, and \textit{Pivoting} as the evaluation tasks, since the other representations cannot handle deformable object manipulation tasks like cloth folding. 
In particular, 3D keypoints~\cite{rekep, iker} rely on fixed relative positions to the object's 6D pose, making them unsuitable for deformable or articulated objects whose spatial structure changes during manipulation. \par 

\noindent \textbf{Key Findings: }Beyond enabling deformable and articulated objects manipulation, \name achieves superior or competitive performance across all assessed tasks in Fig.~\ref{fig:result_rep}, particularly excelling in the contact-rich  \textit{Pivoting} task.
3D keypoints, which capture only initial and final positions without temporal information, struggle in \textit{Pivoting} as they provide limited expert guidance. In contrast, representations with temporal features, \eg, \textit{6D poses} and \textit{3D keypoint trajectories}, perform better by preserving motion dynamics.
While \textit{3D keypoint trajectories} can accelerate early training on \textit{PickNPlace} due to their spatial awareness, they often face challenges when directly serving as policy inputs, particularly in tasks involving orientation changes, such as \textit{Pouring} and \textit{Pivoting}. These limitations underscore the advantages of $\delta$-flow and our policy input design (Sec.~\ref{sec:algo}). Specifically, our $\delta$-flow extracts spatiotemporal manipulation-centric features from low-dimensional representations, making it especially beneficial for challenging tasks with more complex dynamic motions. \par 

\vspace{-0.2cm}
\subsection{Ablation Study}\label{sec:ablation}
\begin{figure}[t]
    \centering
    \includegraphics[width=0.48\textwidth]{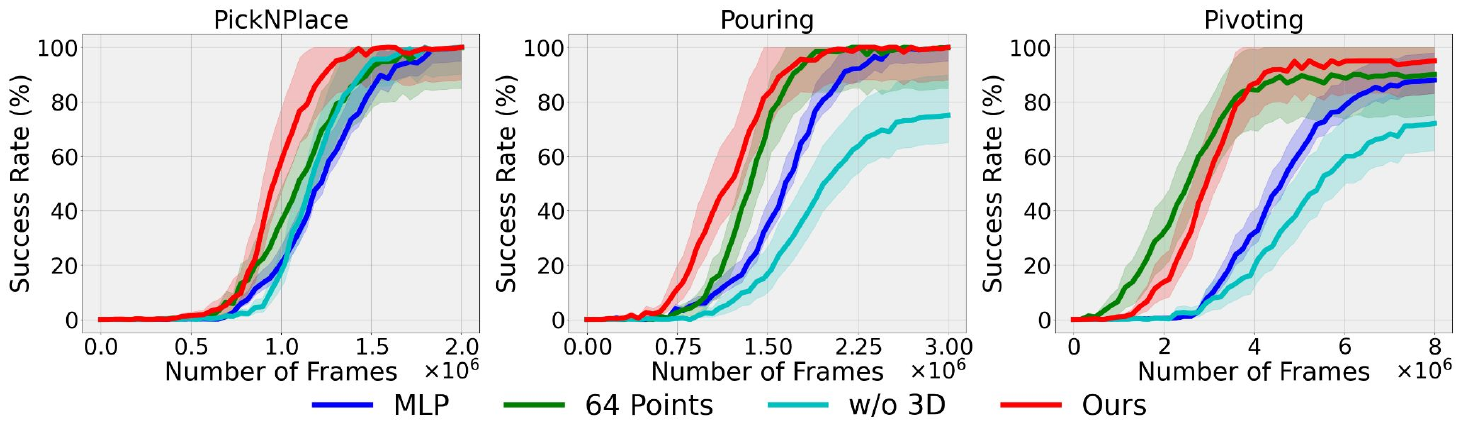}
    \vspace{-0.2in}
    \caption{\textbf{Ablation Evaluations.} The comparison of RL training results with ablations of \name in three tasks. The shaded area represents the standard deviation for three random seeds.}
\label{fig:result_ablation}
\vspace{-0.4cm}
\end{figure}

\vspace{-0.1cm}
We compare the performance of \name with three design variants with controlled experiments on three tasks (\textit{PickNPlace}, \textit{Pouring}, and \textit{Pivoting}): 
\texttt{(1)} \textit{MLP}: Instead of $\delta$-flow, MLP is adopted to encode tracked keypoints into low dimensional space as the policy input. \texttt{(2)} \textit{w/o 3D Initial Centroids~\cite{hudor}}: Removeing initial 3D centroids from \name. 
\texttt{(3)} \textit{different keypoints}: Randomly sampling 64 keypoints instead of 128 keypoints. See our Supp. for more ablations. 
All other settings remain the same and only input representations change. The results are in Fig.~\ref{fig:result_ablation}. \par 
These experiments provide three key insights: \texttt{(a)} Compared to MLP, utilizing observational and future generated $\delta$-flow as inputs better captures the object's spatiotemporal dynamics, leading to improved performance. \texttt{(b)} Incorporating initial 3D keypoints in the input is beneficial since it provides 3D spatial information for learning structured 6D actions. 
\texttt{(c)} Our method is insensitive to the number of keypoints for computing $\delta$-flow. \par 

\subsection{Human Hand Demonstrations on a Real Robot}
\label{sec:real_world}
\begin{figure}[t]
    \centering
    \includegraphics[width=0.48\textwidth]{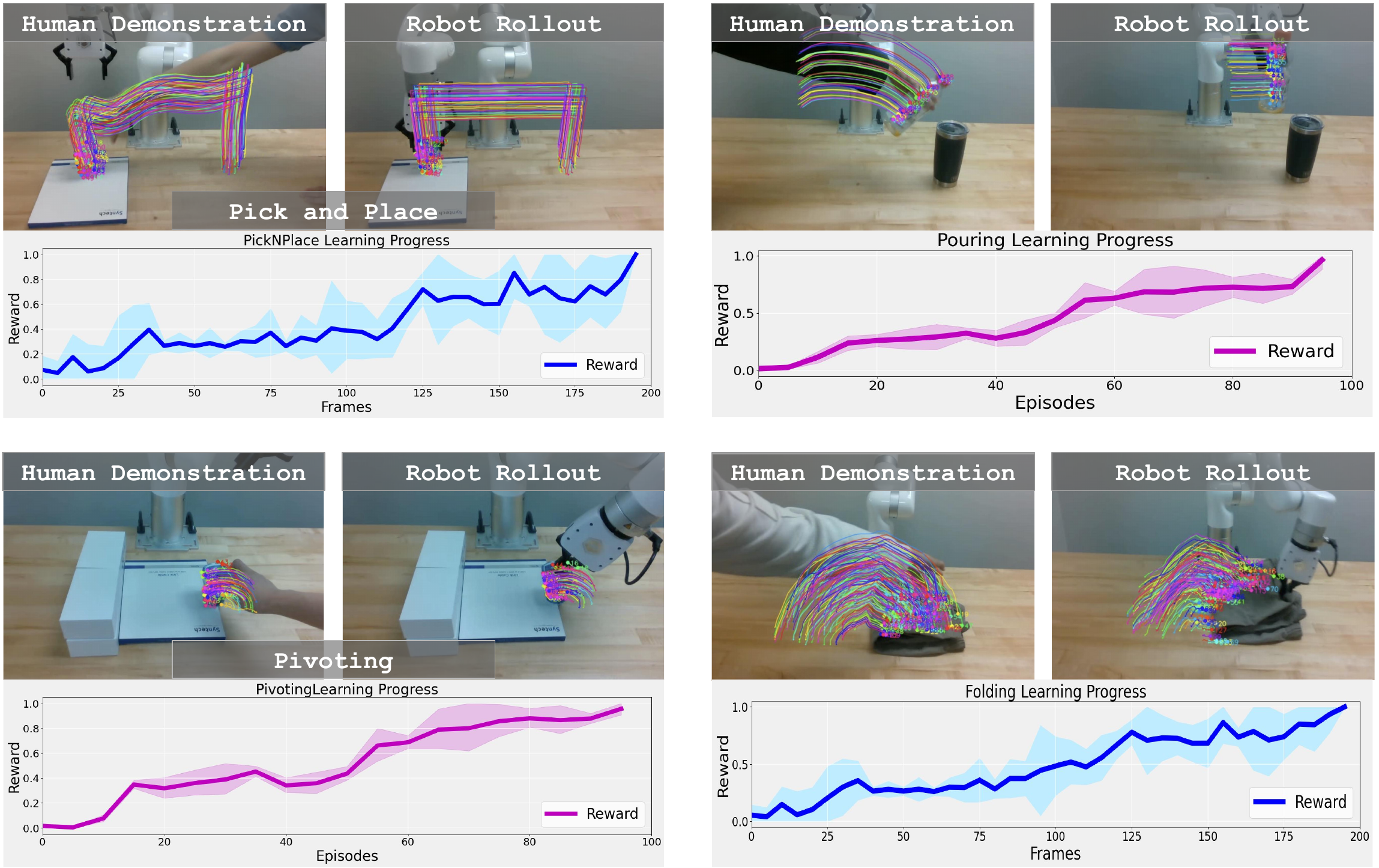}
    \vspace{-0.2cm}
    \caption{\textbf{Real World Evaluations.} The visualization and reward curve of the real world human to robot flow matching. The shade area represents the standard deviation for five random rollouts.}
\label{fig:result_real}
\vspace{-0.1cm}
\end{figure}

In this section, we conduct a case study to show how effectively can our flow-derived reward model perform cross-embodiment human-to-robot transfer. Thus, we set up experiments on four tasks from Im2Flow2Act on a real XArm7 robot arm: \textit{PickNPlace}, \textit{Pouring}, \textbf{Pivoting}, and \textit{Folding}. Our goal is to evaluate whether the cross-embodiment object-centric flow from human collected data can be used for real-robot reward shaping. For each task, we collect five expert human hand demonstrations with varying object placements, and then roll out open-loop robot trajectories aligned with each expert motion. In Fig.~\ref{fig:result_real}, the visualized results confirm that our flow-derived reward is effectively derived from expert object-centric flow from human demonstrations, and produces a monotonic increasing reward signal, indicating the potential of deploying our policy into the real world with easy-to collect human hand demonstrations. We present more implementation details and qualitative results in Appendix.

\vspace{+0.15cm}
\section{Conclusion}
\label{sec:conclusion}
We presented \name, a novel framework that leverages generated flow for reward shaping in visual reinforcement learning, enabling robust and generalizable policy learning. Experiments on 10 challenging manipulation tasks demonstrate the strong performance of our method, outperforming a series of baselines. As future work, we plan to explore whether full 3D flow can overcome the limitations of 2D flow, which may struggle with tasks that involve out-of-plane rotations (\eg, twist-off).

\newpage
{
    \small
    \bibliographystyle{ieeenat_fullname}
    \bibliography{main}

\begin{thebibliography}{59}
\providecommand{\natexlab}[1]{#1}
\providecommand{\url}[1]{\texttt{#1}}
\expandafter\ifx\csname urlstyle\endcsname\relax
  \providecommand{\doi}[1]{doi: #1}\else
  \providecommand{\doi}{doi: \begingroup \urlstyle{rm}\Url}\fi

\bibitem[Burda et~al.(2018)Burda, Edwards, Storkey, and Klimov]{rnd}
Yuri Burda, Harrison Edwards, Amos Storkey, and Oleg Klimov.
\newblock Exploration by random network distillation, 2018.

\bibitem[Chen et~al.(2024)Chen, Wang, Yang, and Liu]{objdex}
Yuanpei Chen, Chen Wang, Yaodong Yang, and C.~Karen Liu.
\newblock Object-centric dexterous manipulation from human motion data, 2024.

\bibitem[Chi et~al.(2024)Chi, Xu, Feng, Cousineau, Du, Burchfiel, Tedrake, and Song]{chi2024diffusionpolicyvisuomotorpolicy}
Cheng Chi, Zhenjia Xu, Siyuan Feng, Eric Cousineau, Yilun Du, Benjamin Burchfiel, Russ Tedrake, and Shuran Song.
\newblock Diffusion policy: Visuomotor policy learning via action diffusion, 2024.

\bibitem[Doersch et~al.(2023)Doersch, Yang, Vecerik, Gokay, Gupta, Aytar, Carreira, and Zisserman]{tapir}
Carl Doersch, Yi Yang, Mel Vecerik, Dilara Gokay, Ankush Gupta, Yusuf Aytar, Joao Carreira, and Andrew Zisserman.
\newblock Tapir: Tracking any point with per-frame initialization and temporal refinement, 2023.

\bibitem[Driess et~al.(2023)Driess, Xia, Sajjadi, Lynch, Chowdhery, Ichter, Wahid, Tompson, Vuong, Yu, Huang, Chebotar, Sermanet, Duckworth, Levine, Vanhoucke, Hausman, Toussaint, Greff, Zeng, Mordatch, and Florence]{palme}
Danny Driess, Fei Xia, Mehdi S.~M. Sajjadi, Corey Lynch, Aakanksha Chowdhery, Brian Ichter, Ayzaan Wahid, Jonathan Tompson, Quan Vuong, Tianhe Yu, Wenlong Huang, Yevgen Chebotar, Pierre Sermanet, Daniel Duckworth, Sergey Levine, Vincent Vanhoucke, Karol Hausman, Marc Toussaint, Klaus Greff, Andy Zeng, Igor Mordatch, and Pete Florence.
\newblock Palm-e: An embodied multimodal language model.
\newblock In \emph{arXiv preprint arXiv:2303.03378}, 2023.

\bibitem[Escontrela et~al.(2024)Escontrela, Adeniji, Yan, Jain, Peng, Goldberg, Lee, Hafner, and Abbeel]{viper}
Alejandro Escontrela, Ademi Adeniji, Wilson Yan, Ajay Jain, Xue~Bin Peng, Ken Goldberg, Youngwoon Lee, Danijar Hafner, and Pieter Abbeel.
\newblock Video prediction models as rewards for reinforcement learning.
\newblock \emph{Advances in Neural Information Processing Systems}, 36, 2024.

\bibitem[Esser et~al.(2020)Esser, Rombach, and Ommer]{vq-gan}
Patrick Esser, Robin Rombach, and Bj{\"{o}}rn Ommer.
\newblock Taming transformers for high-resolution image synthesis.
\newblock \emph{CoRR}, abs/2012.09841, 2020.

\bibitem[Fu et~al.(2018)Fu, Luo, and Levine]{airl}
Justin Fu, Katie Luo, and Sergey Levine.
\newblock Learning robust rewards with adversarial inverse reinforcement learning, 2018.

\bibitem[Gao et~al.(2025)Gao, Zhang, Xu, Zhehao, and Shao]{flip}
Chongkai Gao, Haozhuo Zhang, Zhixuan Xu, Cai Zhehao, and Lin Shao.
\newblock {FLIP}: Flow-centric generative planning for general-purpose manipulation tasks.
\newblock In \emph{The Thirteenth International Conference on Learning Representations}, 2025.

\bibitem[Guo et~al.(2024)Guo, Yang, Rao, Liang, Wang, Qiao, Agrawala, Lin, and Dai]{animatediff}
Yuwei Guo, Ceyuan Yang, Anyi Rao, Zhengyang Liang, Yaohui Wang, Yu Qiao, Maneesh Agrawala, Dahua Lin, and Bo Dai.
\newblock Animatediff: Animate your personalized text-to-image diffusion models without specific tuning, 2024.

\bibitem[Guzey et~al.(2024)Guzey, Dai, Savva, Bhirangi, and Pinto]{hudor}
Irmak Guzey, Yinlong Dai, Georgy Savva, Raunaq Bhirangi, and Lerrel Pinto.
\newblock Bridging the human to robot dexterity gap through object-oriented rewards, 2024.

\bibitem[Haarnoja et~al.(2018)Haarnoja, Zhou, Abbeel, and Levine]{haarnoja2018softactorcriticoffpolicymaximum}
Tuomas Haarnoja, Aurick Zhou, Pieter Abbeel, and Sergey Levine.
\newblock Soft actor-critic: Off-policy maximum entropy deep reinforcement learning with a stochastic actor, 2018.

\bibitem[Han et~al.(2024)Han, Chen, Williams, and Ravichandar]{cimer}
Yunhai Han, Zhenyang Chen, Kyle~A Williams, and Harish Ravichandar.
\newblock Learning prehensile dexterity by imitating and emulating state-only observations, 2024.

\bibitem[Ho and Ermon(2016)]{gail}
Jonathan Ho and Stefano Ermon.
\newblock Generative adversarial imitation learning, 2016.

\bibitem[Ho et~al.(2022)Ho, Salimans, Gritsenko, Chan, Norouzi, and Fleet]{vdm}
Jonathan Ho, Tim Salimans, Alexey Gritsenko, William Chan, Mohammad Norouzi, and David~J Fleet.
\newblock Video diffusion models.
\newblock \emph{Advances in Neural Information Processing Systems}, 35:\penalty0 8633--8646, 2022.

\bibitem[Hsu et~al.(2024)Hsu, Wen, Xu, Narang, Wang, Zhu, Biswas, and Birchfield]{pose}
Cheng-Chun Hsu, Bowen Wen, Jie Xu, Yashraj Narang, Xiaolong Wang, Yuke Zhu, Joydeep Biswas, and Stan Birchfield.
\newblock Spot: Se(3) pose trajectory diffusion for object-centric manipulation, 2024.

\bibitem[Hu et~al.(2021)Hu, Shen, Wallis, Allen-Zhu, Li, Wang, Wang, and Chen]{lora}
Edward~J. Hu, Yelong Shen, Phillip Wallis, Zeyuan Allen-Zhu, Yuanzhi Li, Shean Wang, Lu Wang, and Weizhu Chen.
\newblock Lora: Low-rank adaptation of large language models, 2021.

\bibitem[Huang et~al.(2024{\natexlab{a}})Huang, Levy, Jiang, Anandkumar, Zhu, Fan, Huang, and Shrivastava]{ardup}
Shuaiyi Huang, Mara Levy, Zhenyu Jiang, Anima Anandkumar, Yuke Zhu, Linxi Fan, De-An Huang, and Abhinav Shrivastava.
\newblock Ardup: Active region video diffusion for universal policies.
\newblock In \emph{2024 IEEE/RSJ International Conference on Intelligent Robots and Systems (IROS)}, pages 8465--8472, 2024{\natexlab{a}}.

\bibitem[Huang et~al.(2024{\natexlab{b}})Huang, Jiang, Ze, and Xu]{Huang2023DiffusionReward}
Tao Huang, Guangqi Jiang, Yanjie Ze, and Huazhe Xu.
\newblock Diffusion reward: Learning rewards via conditional video diffusion.
\newblock \emph{European Conference on Computer Vision (ECCV)}, 2024{\natexlab{b}}.

\bibitem[Huang et~al.(2024{\natexlab{c}})Huang, Wang, Li, Zhang, and Fei-Fei]{rekep}
Wenlong Huang, Chen Wang, Yunzhu Li, Ruohan Zhang, and Li Fei-Fei.
\newblock Rekep: Spatio-temporal reasoning of relational keypoint constraints for robotic manipulation.
\newblock In \emph{8th Annual Conference on Robot Learning}, 2024{\natexlab{c}}.

\bibitem[Jiang et~al.(2024)Jiang, Sun, Huang, Li, Liang, and Xu]{MCR}
Guangqi Jiang, Yifei Sun, Tao Huang, Huanyu Li, Yongyuan Liang, and Huazhe Xu.
\newblock Robots pre-train robots: Manipulation-centric robotic representation from large-scale robot datasets.
\newblock \emph{arXiv preprint arXiv:2410.22325}, 2024.

\bibitem[Karaev et~al.(2023)Karaev, Rocco, Graham, Neverova, Vedaldi, and Rupprecht]{cotracker}
Nikita Karaev, Ignacio Rocco, Benjamin Graham, Natalia Neverova, Andrea Vedaldi, and Christian Rupprecht.
\newblock {CoTracker}: It is better to track together.
\newblock 2023.

\bibitem[Kareer et~al.(2024)Kareer, Patel, Punamiya, Mathur, Cheng, Wang, Hoffman, and Xu]{egomimic}
Simar Kareer, Dhruv Patel, Ryan Punamiya, Pranay Mathur, Shuo Cheng, Chen Wang, Judy Hoffman, and Danfei Xu.
\newblock Egomimic: Scaling imitation learning via egocentric video, 2024.

\bibitem[Kirillov et~al.(2023)Kirillov, Mintun, Ravi, Mao, Rolland, Gustafon, Xiao, Whitehead, Berg, Lo, Doll{\'a}r, and Girshick]{sam}
Alexander Kirillov, Eric Mintun, Nikhila Ravi, Hanzi Mao, Paul Rolland, Laura Gustafon, Tete Xiao, Spencer Whitehead, Alexander~C. Berg, Wan-Yen Lo, Piotr Doll{\'a}r, and Ross Girshick.
\newblock Segment anything.
\newblock \emph{arXiv preprint arXiv:2304.02643}, 2023.

\bibitem[Ko et~al.(2023)Ko, Mao, Du, Sun, and Tenenbaum]{avdc}
Po-Chen Ko, Jiayuan Mao, Yilun Du, Shao-Hua Sun, and Joshua~B Tenenbaum.
\newblock {Learning to Act from Actionless Videos through Dense Correspondences}.
\newblock \emph{arXiv:2310.08576}, 2023.

\bibitem[Levine et~al.(2016)Levine, Pastor, Krizhevsky, et~al.]{robot-data-collection}
S Levine, P Pastor, A Krizhevsky, et~al.
\newblock Learning hand-eye coordination for robotic grasping with large-scale data collection. int symp on experimental robotics, 2016.

\bibitem[Levine et~al.(2020)Levine, Kumar, Tucker, and Fu]{viper-24}
Sergey Levine, Aviral Kumar, George Tucker, and Justin Fu.
\newblock Offline reinforcement learning: Tutorial, review, and perspectives on open problems.
\newblock \emph{arXiv preprint arXiv:2005.01643}, 2020.

\bibitem[Levy et~al.(2024)Levy, Haldar, Pinto, and Shirivastava]{point}
Mara Levy, Siddhant Haldar, Lerrel Pinto, and Abhinav Shirivastava.
\newblock P3-po: Prescriptive point priors for visuo-spatial generalization of robot policies, 2024.

\bibitem[Li et~al.(2024)Li, Yu, Cheng, and Xu]{league}
Zhaoyi Li, Kelin Yu, Shuo Cheng, and Danfei Xu.
\newblock {LEAGUE}++: {EMPOWERING} {CONTINUAL} {ROBOT} {LEARNING} {THROUGH} {GUIDED} {SKILL} {ACQUISITION} {WITH} {LARGE} {LANGUAGE} {MODELS}.
\newblock In \emph{ICLR 2024 Workshop on Large Language Model (LLM) Agents}, 2024.

\bibitem[Liang et~al.(2022)Liang, Huang, Xia, Xu, Hausman, Ichter, Florence, and Zeng]{codeaspolicy}
Jacky Liang, Wenlong Huang, Fei Xia, Peng Xu, Karol Hausman, Brian Ichter, Pete Florence, and Andy Zeng.
\newblock Code as policies: Language model programs for embodied control.
\newblock In \emph{arXiv preprint arXiv:2209.07753}, 2022.

\bibitem[Liang et~al.(2024)Liang, Liu, Ozguroglu, Sudhakar, Dave, Tokmakov, Song, and Vondrick]{dreamite}
Junbang Liang, Ruoshi Liu, Ege Ozguroglu, Sruthi Sudhakar, Achal Dave, Pavel Tokmakov, Shuran Song, and Carl Vondrick.
\newblock Dreamitate: Real-world visuomotor policy learning via video generation.
\newblock \emph{arXiv preprint arXiv:2406.16862}, 2024.

\bibitem[Lin et~al.(2024)Lin, Cui, Xie, Hua, and Sadigh]{flowretrieval}
Li-Heng Lin, Yuchen Cui, Amber Xie, Tianyu Hua, and Dorsa Sadigh.
\newblock Flowretrieval: Flow-guided data retrieval for few-shot imitation learning.
\newblock In \emph{8th Annual Conference on Robot Learning}, 2024.

\bibitem[Liu et~al.(2025)Liu, Zeng, Ren, Li, Zhang, Yang, Jiang, Li, Yang, Su, et~al.]{grounding-dino}
Shilong Liu, Zhaoyang Zeng, Tianhe Ren, Feng Li, Hao Zhang, Jie Yang, Qing Jiang, Chunyuan Li, Jianwei Yang, Hang Su, et~al.
\newblock Grounding dino: Marrying dino with grounded pre-training for open-set object detection.
\newblock In \emph{European Conference on Computer Vision}, pages 38--55. Springer, 2025.

\bibitem[Loshchilov and Hutter(2019)]{adamw}
Ilya Loshchilov and Frank Hutter.
\newblock Decoupled weight decay regularization, 2019.

\bibitem[Ma et~al.(2023{\natexlab{a}})Ma, Liang, Som, Kumar, Zhang, Bastani, and Jayaraman]{LIV}
Yecheng~Jason Ma, William Liang, Vaidehi Som, Vikash Kumar, Amy Zhang, Osbert Bastani, and Dinesh Jayaraman.
\newblock Liv: Language-image representations and rewards for robotic control, 2023{\natexlab{a}}.

\bibitem[Ma et~al.(2023{\natexlab{b}})Ma, Liang, Wang, Huang, Bastani, Jayaraman, Zhu, Fan, and Anandkumar]{eureka}
Yecheng~Jason Ma, William Liang, Guanzhi Wang, De-An Huang, Osbert Bastani, Dinesh Jayaraman, Yuke Zhu, Linxi Fan, and Anima Anandkumar.
\newblock Eureka: Human-level reward design via coding large language models.
\newblock \emph{arXiv preprint arXiv: Arxiv-2310.12931}, 2023{\natexlab{b}}.

\bibitem[Patel et~al.(2025)Patel, Yin, Huang, Garg, Nayyeri, Fei-Fei, Lazebnik, and Li]{iker}
Shivansh Patel, Xinchen Yin, Wenlong Huang, Shubham Garg, Hooshang Nayyeri, Li Fei-Fei, Svetlana Lazebnik, and Yunzhu Li.
\newblock A real-to-sim-to-real approach to robotic manipulation with vlm-generated iterative keypoint rewards, 2025.

\bibitem[Radford et~al.(2021)Radford, Kim, Hallacy, Ramesh, Goh, Agarwal, Sastry, Askell, Mishkin, Clark, Krueger, and Sutskever]{clip}
Alec Radford, Jong~Wook Kim, Chris Hallacy, Aditya Ramesh, Gabriel Goh, Sandhini Agarwal, Girish Sastry, Amanda Askell, Pamela Mishkin, Jack Clark, Gretchen Krueger, and Ilya Sutskever.
\newblock Learning transferable visual models from natural language supervision, 2021.

\bibitem[Rombach et~al.(2022)Rombach, Blattmann, Lorenz, Esser, and Ommer]{stablediffusion}
Robin Rombach, Andreas Blattmann, Dominik Lorenz, Patrick Esser, and Björn Ommer.
\newblock High-resolution image synthesis with latent diffusion models, 2022.

\bibitem[Schulman et~al.(2017)Schulman, Wolski, Dhariwal, Radford, and Klimov]{schulman2017proximalpolicyoptimizationalgorithms}
John Schulman, Filip Wolski, Prafulla Dhariwal, Alec Radford, and Oleg Klimov.
\newblock Proximal policy optimization algorithms, 2017.

\bibitem[Su et~al.(2025)Su, Zhan, Fang, Li, Lu, and Yang]{MBA}
Yue Su, Xinyu Zhan, Hongjie Fang, Yong-Lu Li, Cewu Lu, and Lixin Yang.
\newblock Motion before action: Diffusing object motion as manipulation condition, 2025.

\bibitem[Tang et~al.(2024)Tang, Pan, Zhan, Zhou, Yao, Liu, Tomizuka, Ding, and Fu]{part}
Weiliang Tang, Jia-Hui Pan, Wei Zhan, Jianshu Zhou, Huaxiu Yao, Yun-Hui Liu, Masayoshi Tomizuka, Mingyu Ding, and Chi-Wing Fu.
\newblock Embodiment-agnostic action planning via object-part scene flow, 2024.

\bibitem[Team(2024)]{gpt4}
OpenAI GPT-4 Team.
\newblock Gpt-4 technical report, 2024.

\bibitem[Todorov et~al.(2012)Todorov, Erez, and Tassa]{mujoco}
Emanuel Todorov, Tom Erez, and Yuval Tassa.
\newblock Mujoco: A physics engine for model-based control.
\newblock In \emph{2012 IEEE/RSJ International Conference on Intelligent Robots and Systems}, pages 5026--5033, 2012.

\bibitem[Touvron et~al.(2023)Touvron, Lavril, Izacard, Martinet, Lachaux, Lacroix, Rozière, Goyal, Hambro, Azhar, Rodriguez, Joulin, Grave, and Lample]{llama}
Hugo Touvron, Thibaut Lavril, Gautier Izacard, Xavier Martinet, Marie-Anne Lachaux, Timothée Lacroix, Baptiste Rozière, Naman Goyal, Eric Hambro, Faisal Azhar, Aurelien Rodriguez, Armand Joulin, Edouard Grave, and Guillaume Lample.
\newblock Llama: Open and efficient foundation language models, 2023.

\bibitem[Wang et~al.(2024{\natexlab{a}})Wang, Sridhar, Feng, Van~der Merwe, et~al.]{this-that}
Boyang Wang, Nikhil Sridhar, Chao Feng, Mark Van~der Merwe, et~al.
\newblock This\&that: Language-gesture controlled video generation for robot planning.
\newblock \emph{arXiv preprint arXiv:2407.05530}, 2024{\natexlab{a}}.

\bibitem[Wang et~al.(2024{\natexlab{b}})Wang, Sun, Zhang, Xian, Biyik, Held, and Erickson]{rlvlm}
Yufei Wang, Zhanyi Sun, Jesse Zhang, Zhou Xian, Erdem Biyik, David Held, and Zackory Erickson.
\newblock Rl-vlm-f: Reinforcement learning from vision language foundation model feedback, 2024{\natexlab{b}}.

\bibitem[Wu et~al.(2024)Wu, Yin, Feng, He, Li, Hao, and Long]{ivideogpt}
Jialong Wu, Shaofeng Yin, Ningya Feng, Xu He, Dong Li, Jianye Hao, and Mingsheng Long.
\newblock ivideogpt: Interactive videogpts are scalable world models.
\newblock \emph{Advances in Neural Information Processing Systems}, 37:\penalty0 68082--68119, 2024.

\bibitem[Xie et~al.(2024)Xie, Zhao, Wu, Liu, Luo, Zhong, Yang, and Yu]{text2reward}
Tianbao Xie, Siheng Zhao, Chen~Henry Wu, Yitao Liu, Qian Luo, Victor Zhong, Yanchao Yang, and Tao Yu.
\newblock Text2reward: Reward shaping with language models for reinforcement learning, 2024.

\bibitem[Xiong et~al.(2021)Xiong, Li, Chen, Bharadhwaj, Sinha, and Garg]{watching}
Haoyu Xiong, Quanzhou Li, Yun-Chun Chen, Homanga Bharadhwaj, Samarth Sinha, and Animesh Garg.
\newblock Learning by watching: Physical imitation of manipulation skills from human videos.
\newblock In \emph{2021 IEEE/RSJ International Conference on Intelligent Robots and Systems (IROS)}, pages 7827--7834. IEEE, 2021.

\bibitem[Xu et~al.(2024)Xu, Xu, Xu, Chi, Wetzstein, Veloso, and Song]{flow2act}
Mengda Xu, Zhenjia Xu, Yinghao Xu, Cheng Chi, Gordon Wetzstein, Manuela Veloso, and Shuran Song.
\newblock Flow as the cross-domain manipulation interface, 2024.

\bibitem[Yan et~al.(2021)Yan, Zhang, Abbeel, and Srinivas]{videogpt}
Wilson Yan, Yunzhi Zhang, Pieter Abbeel, and Aravind Srinivas.
\newblock Videogpt: Video generation using vq-vae and transformers, 2021.

\bibitem[Yang et~al.(2024)Yang, Tjia, Berg, Damen, Agrawal, and Gupta]{rank2reward}
Daniel Yang, Davin Tjia, Jacob Berg, Dima Damen, Pulkit Agrawal, and Abhishek Gupta.
\newblock Rank2reward: Learning shaped reward functions from passive video.
\newblock In \emph{2024 IEEE International Conference on Robotics and Automation (ICRA)}, pages 2806--2813. IEEE, 2024.

\bibitem[Yarats et~al.(2022)Yarats, Fergus, Lazaric, and Pinto]{yarats2022mastering}
Denis Yarats, Rob Fergus, Alessandro Lazaric, and Lerrel Pinto.
\newblock Mastering visual continuous control: Improved data-augmented reinforcement learning.
\newblock In \emph{International Conference on Learning Representations}, 2022.

\bibitem[Yu et~al.(2024)Yu, Han, Wang, Saxena, Xu, and Zhao]{mimictouch}
Kelin Yu, Yunhai Han, Qixian Wang, Vaibhav Saxena, Danfei Xu, and Ye Zhao.
\newblock Mimictouch: Leveraging multi-modal human tactile demonstrations for contact-rich manipulation.
\newblock In \emph{8th Annual Conference on Robot Learning}, 2024.

\bibitem[Yu et~al.(2025)Yu, Bhaskar, Singh, Mahammad, and Tokekar]{sketch}
Peihong Yu, Amisha Bhaskar, Anukriti Singh, Zahiruddin Mahammad, and Pratap Tokekar.
\newblock Sketch-to-skill: Bootstrapping robot learning with human drawn trajectory sketches, 2025.

\bibitem[Yu et~al.(2021)Yu, Quillen, He, Julian, Narayan, Shively, Bellathur, Hausman, Finn, and Levine]{metaworld}
Tianhe Yu, Deirdre Quillen, Zhanpeng He, Ryan Julian, Avnish Narayan, Hayden Shively, Adithya Bellathur, Karol Hausman, Chelsea Finn, and Sergey Levine.
\newblock Meta-world: A benchmark and evaluation for multi-task and meta reinforcement learning, 2021.

\bibitem[Yuan et~al.(2024)Yuan, Wen, Zhang, and Gao]{flow-15}
Chengbo Yuan, Chuan Wen, Tong Zhang, and Yang Gao.
\newblock General flow as foundation affordance for scalable robot learning.
\newblock \emph{arXiv preprint arXiv:2401.11439}, 2024.

\bibitem[Zheng et~al.(2024)Zheng, Liang, Huang, Gao, Daum{\'e}~III, Kolobov, Huang, and Yang]{trace}
Ruijie Zheng, Yongyuan Liang, Shuaiyi Huang, Jianfeng Gao, Hal Daum{\'e}~III, Andrey Kolobov, Furong Huang, and Jianwei Yang.
\newblock Tracevla: Visual trace prompting enhances spatial-temporal awareness for generalist robotic policies.
\newblock \emph{arXiv preprint arXiv:2412.10345}, 2024.

\end{thebibliography}
}

\clearpage
\let\cleardoublepage\clearpage
\appendix
\section{Cross-Embodiment and Cross-Domain Data Collection}
To train the flow generation model, we collect a dataset contains $12k$ trajectories of three different embodiments, where they work with ten different tasks from two different task domains. We aim to utilize different embodiments for highlighting that object-centric flow can be trained with large scale diverse training data. \par 

\begin{figure}[t]
\vspace{-0.2cm}
    \centering
    \includegraphics[width=0.5\textwidth]{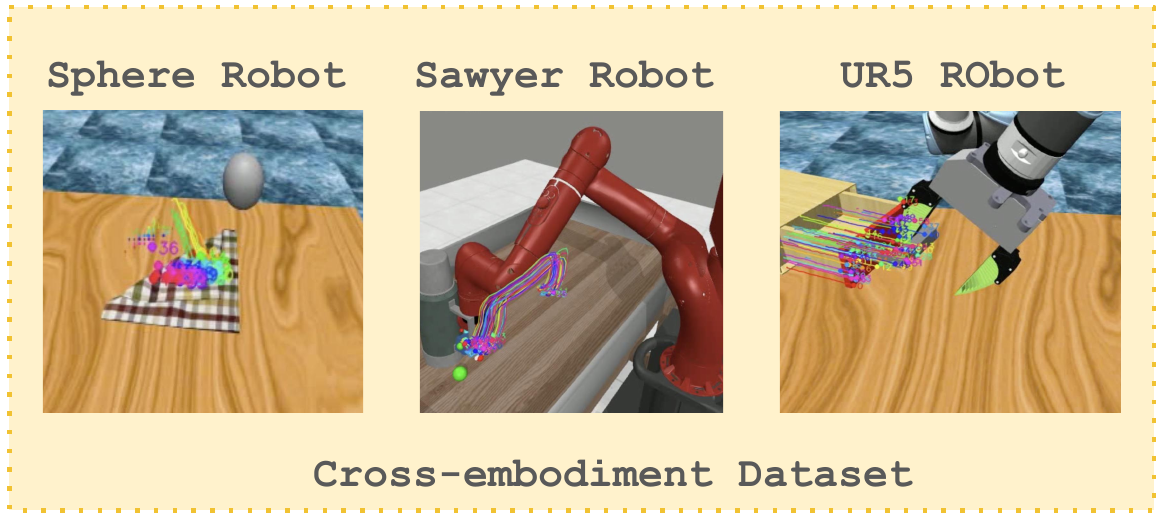}
    \vspace{-0.4cm}
    \caption{\textbf{Visualization of cross-embodiment data.}}
\label{fig:colleciton}
\vspace{-0.4cm}
\end{figure}

In the setting of Im2Flow2Act~\cite{flow2act}, we utilize their sphere robot dataset as the first kind of cross-embodiment data. Four tasks, which are \textit{PickNPlace}, \textit{Pouring}, \textit{Opening}, and \textit{Folding}, are used for data collection. Also shown in Im2Flow2Act~\cite{flow2act}, this kind of data are proposed to emulate the cross-embodiment human data in the real world. Out of those tasks, we design a new contact-rich manipulation task \textit{Pivoting} with the UR5 robot for collecting data. To collect robot data, we place the robot in different initial positions and use a manipulation script to move it to five different contact points. Then, we apply five different action scripts to enable the robot to stand the peg up and make contact with the wall. For data in the MetaWorld~\cite{metaworld}, the task setting and robot used for data collection was totally different from the Im2Flow2Act\cite{flow2act}. We choose five different tasks, which are \textit{Assembly}, \textit{Coffee Push}, \textit{Door Close}, \textit{Lever Pull}, and \textit{Stick Pull}. Those data are collected by the Sawyer Robot. In this benchmark, we rollout the trained RL model in MetaWorld~\cite{metaworld} for data collection. Each task contain 1200 trajectories, where we have 12k training data in total. The visualization of each embodiment are shown in Fig.~\ref{fig:colleciton}. \par

\section{Tasks Descriptions}
We select four tasks from Im2Flow2Act~\cite{flow2act}, five tasks from MetaWorld~\cite{metaworld}, and one self-designed task \textit{pivoting}.
\begin{itemize}
    \item \textit{PickNPlace: }Pick a mug from one random position to the bowl in another random position.
    \item \textit{Pouring: }Pick a mug from one random position and pour the water to a bowl in another random position.
    \item \textit{Opening: }Grasp the handle and open the cabinet.
    \item \textit{Folding: }Fold the cloth from one corner to another corner.
    \item \textit{Pivoting: }Contact with the peg without grasping, and pivot it to stand up by interacting with the wall.
    \item \textit{Coffee Push: }Push the coffee mug to a specific position.
    \item \textit{Door Close: }Close the door of a cabinet.
    \item \textit{Assembly: }Pick up a stick and align its square hole with a peg on the table.
    \item \textit{Lever Pull: }Grasp the lever and pull it up to the up right position.
    \item \textit{Stick Pull:} Pick up a stick, insert it into a kettle, and pull the kettle to a specific position.
\end{itemize}

\section{Flow Generation Implementation Details}
In this section, we aim to share more details about implementation, training, and processing of our flow generation model, which is similar to Im2Flow2Act~\cite{flow2act}.
\subsection{Implementations }
The first step is getting the training dataset. We use Grounding DINO~\cite{grounding-dino} to detect the bounding box of the described object from the initial RGB frame, and then uniformly sample keypoints within the box. To track those keypoints from the video, we apply the SOTA keypoint tracking foundation model~\cite{cotracker} for keypoint tracking. Unlike the TAPIR~\cite{tapir} used in previous work~\cite{flow2act}, CoTracker~\cite{cotracker} can track occluded objects in the image, which is extremely important for contact-rich manipulation tasks.  Then, we formulate the tracked keypoints as object-centric flow $\mathcal{F}_0 \in \mathbb{R}^{3 \times T \times H \times W}$ with temporal representation in T time space. The first two channels represent the pixel coordinates of object keypoints in image space, while the third represents their visibility during the execution. \par 

The generated flow is conditioned on the initial image of the task, the initial keypoints, and the text description. The encoder deisgn for the inputs are also the same as the ~\cite{flow2act}. The text descriptions are processed into the CIIP~\cite{clip} to obtain text embeddings. For the initial image, we utilize the CLIP encoder to get the patch embeddings. The initial keypoints are encoded through fixed 2D sinusoidal positional encoding. Finally, those inputs are processed into the denosing process  through cross-attention. \par 

With this flow representation, we can leverage the diffusion-based video generation based on AnimateDiff [18] for flow generation. Same as the Im2FLow2Act~\cite{flow2act}, we encode the object flow into a latent space and train the generative model based on it. Similar to the StableDiffusion~\cite{stablediffusion},  we use the auto encoder VA-GAN~\cite{vq-gan} to encode the flow into low dimentional embeddings. Then, to utilize the low-dimensional latent space, we utilize a two-stage training process. Firstly, we fix the encoder from the AE and finetune the pretrained decoder to better adapt it to the flow images. Then, same as the Im2Flow2Act, we insert the motion module layer into StableDiffusion proposed by Animatediff~\cite{animatediff} to model the temporal dynamics for flow generation. The second stage is training the motion module layer from scratch but only insert LoRA (Low-Rank Adaptation) layers~\cite{lora} into the SD model. \par 

\subsection{Training Details}
Training with cross-embodiment data from two different task domains, we process both the image from the tasks in Im2Flow2Act~\cite{flow2act} and MetaWorld~\cite{metaworld} into resolutions $480 \times 480$. For extracting keypoints from the bounding box generated by Grounding Dino~\cite{grounding-dino}, we set the spatial and temporal resolution to H = W = 32 and T = 32 for generating flow for 1024 keypoints over 100 steps, which also means that 100 keypoints set can be used for reward shaping in our reward model. To train the model on our cross-embodiment dataset, we firstly train the decoder of VQ-GAN~\cite{vq-gan} in StableDiffusion~\cite{stablediffusion} for 400 epoches with a learning rate of $5e-5$. Secondly, for training AnimateDiff, we insert the LoRA~\cite{lora} with a rank of 128 into the Unet from StableDiffusion and train the motion module layer from scratch with the same hyperparameter shown in Im2Flow2Act~\cite{flow2act}, which is trained with learning rate of $1e-4$ for 300 epochs using Adamw~\cite{adamw} optimizer with weight deacy $1e-2$ betas $(0.9, 0.999),$ and epsilon $1e-8$. \par 

\subsection{Flow Post Processing}
Similar to Im2Flow2Act~\cite{flow2act}, we use motion filter to extract moving keypoints from the object itself. More specifically, we use moving filter to remove those static keypoints and use SAM~\cite{sam} filter to remove keypoints which are not on the object, such as keypoints on the robot or the table. \par 

\textbf{Moving Filter: } Since some of the keypoints selected from bounding box are sampled from the environment, we use the moving filter to extract the moving points from those keypoints. Then, we use moving filter to remove those keypoints whose movement in the image space ($480 \times 480$) is below a certain threshold. For all the tasks, we select 50 pixels as the threshold for removing those static keypoints. This method can effectively remove those background keypoints. \par 

\textbf{SAM Filter~\cite{sam}: }Since we also use robot data in our dataset, those keypoints on the robot are also will be counted as moving keypoints. Then, using SAM~\cite{sam} to do semantic segmentation and remove those moving keypoints on the robot is necessary. We utilize SAM to obtain the segmentation and then iterate through the keypoints, filtering out those whose corresponding segment area exceeds a predefined threshold. To preserve keypoints on objects with rich textures, we set a high threshold of 10,000 across all tasks.
Finally, we randomly sampled 128 points from the selected keypoints for our reward model and policy input. \par 

\section{RL Implementation Details }
\subsection{Reward and Policy Input with Generated Flow}
For both RL training in Im2Flow2Act~\cite{flow2act} and MetaWorld~\cite{metaworld}, we firstly need to generate the initial flow from the first frame for each iteration, which will be used for both policy input and reward shaping. Then, the observation space of the policy input is $128 \times 3$, where we sampled 128 keypoints from the object. \par 

To do online flow matching with the generated flow, we also utilize CoTracker~\cite{cotracker} for online tracking in the real-robot execution. Using the same motion filters, 128 keypoints will be sampled from the realtime motion. The online keypoints tracking will be generated for each timestep during the robot execution. Then, those keypoints will be processed as $\delta$-flow for reward shaping. \par 

Since our flow generation model will generate 100 frames with keypoints subset, we can use each centroid calculated from the keypoints subset of those 100 frames as reward. For reward calculation and policy input. We limit the $max\_step\_episode$ into number less or equal to 100 steps (respectively for different tasks) and use them for real-time flow matching in reward generation and policy learning. \par 

\subsection{RL Reward Design}
Out of the flow-derived reward, our RL training Pipeline also requires specific reward design for achieving the goal.

\textbf{Reaching Reward: }For all the tasks, we need to define similar reaching reward to guide to robot move toward the object. For tasks which required robot to grasp the object, robot needs to open their gripper and reach the grasping position. For the tasks which required robot to push or contact with the object, the robot will be guided to move to contact with a certain area. The reward will be define as: $\left( 1 - \tanh(10.0 \cdot d_{\text{grip}}) \right)$.

\textbf{Grasping / Contact Reward: }We also design sparse reward as a subgoal for guiding robot to accomplish the task. For grasping task, we will set the reward to be $0.25$ as the reward. For contact reward, we will set the reward to be $0.25$ once the gripper is contact with a certain area of the object.

\textbf{Goal-conditioned Reward: }For all the task, we need to define a goal state to showcase that the robot successfully acchieve the goal. For most of the task, it should be easily, such as those defined task in MetaWorld~\cite{metaworld} and some simple task like \textit{PickNPlace}. We can just define the final position for object to be. For some other harder-to-define tasks like pouring, we set up a target pose range as the goal, which is limited to a certain position with certain orientations, where the orientation is sampled from $(\frac{5\pi}{16}\frac{7\pi}{16})$. For opening, the reward will be defined by the opening distance, which is $0.1m$. 

\subsection{Training Details}
The Training hyperper parameters have been shown in Table~\ref{tab:parameter}.

\begin{table}[h]
    \centering
    \caption{Hyperparameters for DrQv2 with Flow-derived Reward.}
    \begin{tabular}{l|l}
        \toprule
        \textbf{Hyperparameter} & \textbf{Value} \\
        \midrule
        \multicolumn{2}{l}{\textbf{Environment}} \\
        \midrule
        Action repeat & 3 (MetaWorld) \\
                     & 3 (Im2Flow2Act) \\
        Frame stack & 1 \\
        Rendered Image & $480 \times 480$ \\
        Observation size & $128 \times 3$ \\
        Reward type & Sparse \\
        \midrule
        \multicolumn{2}{l}{\textbf{DrQv2}} \\
        \midrule
        Data Augmentation & $\pm 4$ RandomShift \\
        Replay buffer capacity & $10^6$ \\
        Discount $\gamma$ & 0.99 \\
        $n$-step returns & 3 \\
        Seed frames & 4000 \\
        Exploration steps & 2000 \\
        Exploration stddev. clip & 0.3 \\
        Exploration stddev. schedule & Linear(1.0, 0.1, $3 \times 10^6$) \\
        Soft update rate & 0.01 \\
        Optimizer & Adam \\
        Batch size & 256 \\
        Update frequency & 2 \\
        Learning rate & $10^{-4}$ \\
        \bottomrule
    \end{tabular}
    \label{tab:parameter}
\end{table}

\subsection{Visualization of delta-flow model}
\begin{figure}[h]
    \centering
    \includegraphics[width=\linewidth]{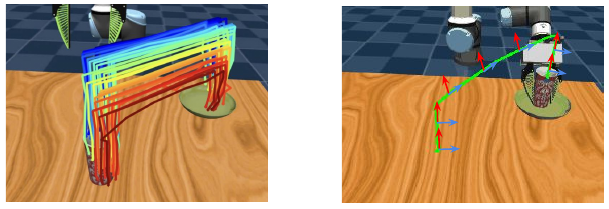}
    \caption{Visualization of the 2d delta-flow extractions.}
    \label{fig:flow}
\end{figure}

To formulate the delta-flow reward model, we first calculate the centroid of the flow at each time step and calculate the relative translation and rotations between the current step and initial step. More visualization of the calculated centroid at each time step is shown in Fig.~\ref{fig:flow}.

\section{Ablation Study}
\subsection{Ablation Study for Robustness of reward model}
\begin{figure}[t]
\vspace{-0.2cm}
    \centering
    \includegraphics[width=0.5\textwidth]{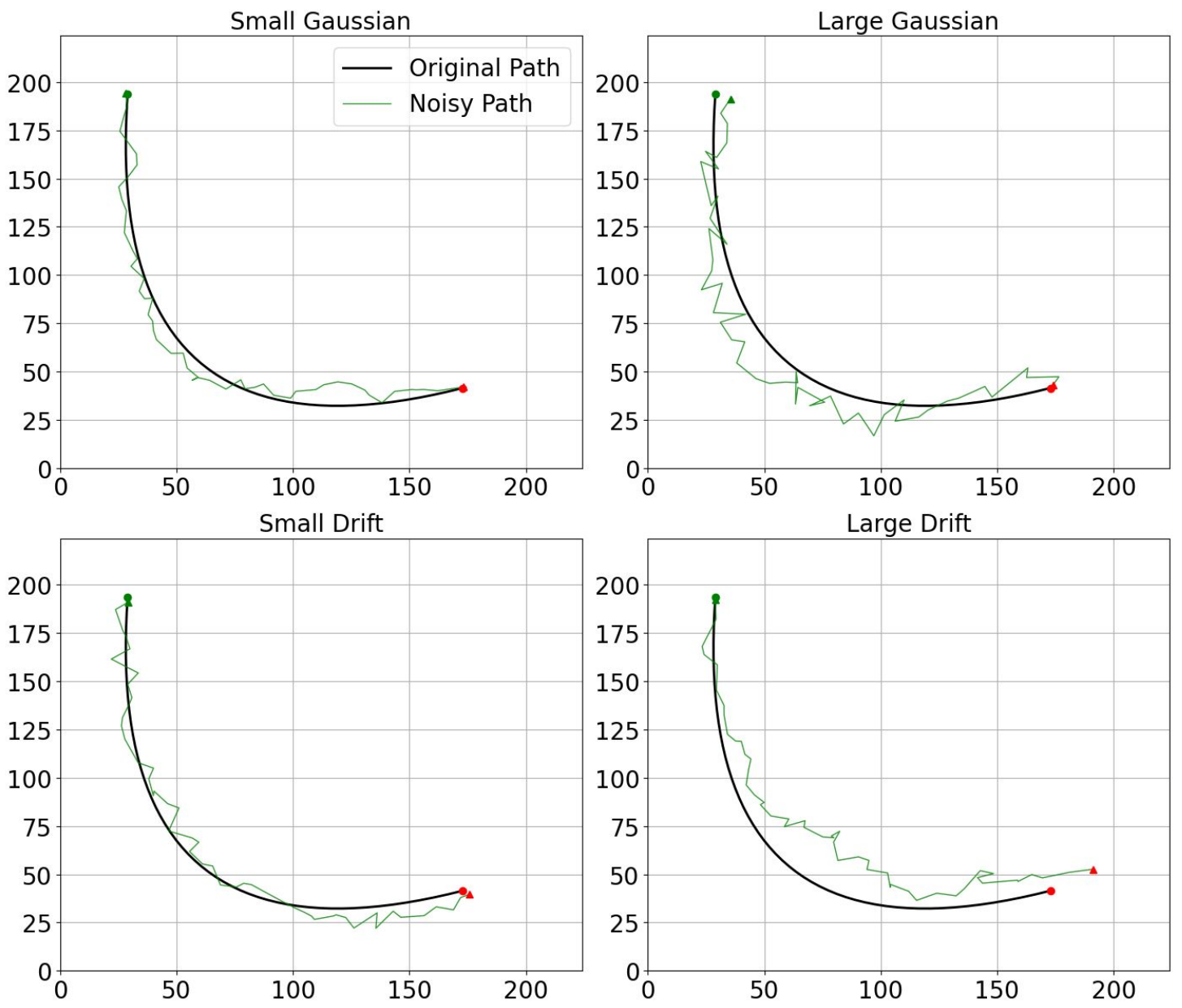}
    \vspace{-0.2cm}
    \caption{Visualization of the simulated noised 2D trajectory.}
\label{fig:noise}
\end{figure}

In real-world scenarios, the flow trajectory predicted by the diffusion model or the tracker can be noisy. To investigate the effect of these noises on policy performance, we simulate real-world noise for our method. Specifically, we build a noise model to cumulatively perturb trajectories and deviate endpoints, using the addition of Brownian motion (\ie, Gaussian random walk) and Brownian bridge (\ie, trajectories with predefined perturbed endpoints), with separate controllable standard deviations. We set up four types of random noise: small Gaussian (gauss=1x, drift=0x), large Gaussian (gauss=4x, drift=0x), small drift (gauss=2x, drift=1x), and large drift (gauss=2x, drift=2x).
A vivid visualization of our noise composite model applied to a Bessel smoothed trajectory is displayed in Fig.~\ref{fig:noise}. 

We evaluate the performance of our model after applying this noise model to the generated flow in five challenging tasks. The result is shown in Table~\ref{tab:noise}. Since our model is already trained on generated flows with different magnitudes of noise and due to our task-oriented rewards, it has the robustness to noise in the generated flows. Therefore, our method still achieves a similar performance with trajectory noise, especially for the cases with large Gaussian noises. 

Furthermore, our method can maintain relatively high performance when the goal position is largely drifted from the ground-truth by $\geq 20$ pixels in the tasks with position-sensitive evaluation, \eg, \texttt{Folding} and \texttt{Pouring} (Fig.~\ref{fig:noise}).

\begin{table}[h]
\centering
\scriptsize
\setlength{\tabcolsep}{8pt}
\begin{tabular}{l|c c c c c}
\toprule
 & PickNP. & Pour & Open & Fold & Pivot \\ \midrule
\name & 95 & 95 & 95 & 80 & 85 \\
+Gauss$\times 1$ Drift$\times 0$ & 95 & 95 & 90 & 80 & 85 \\ 
+Gauss$\times 4$ Drift$\times 0$ & 95 & 90 & 90 & 75 & 80 \\ 
+Gauss$\times 2$ Drift$\times 1$ & 95 & 90 & 90 & 70 & 85 \\ 
+Gauss$\times 2$ Drift$\times 2$ & 85 & 75 & 85 & 65 & 75 \\ \bottomrule
\end{tabular}
\caption{Performance for noise sensitivity analysis on five tasks. Noises are added to flow trajectories for comparisons.}
\label{tab:noise}
\vspace{-0.2cm}
\end{table}

\subsection{Ablation Study of Different Number of Keypoints}
\begin{figure}[h]
\vspace{-0.4cm}
    \centering
    \includegraphics[width=0.9\linewidth]{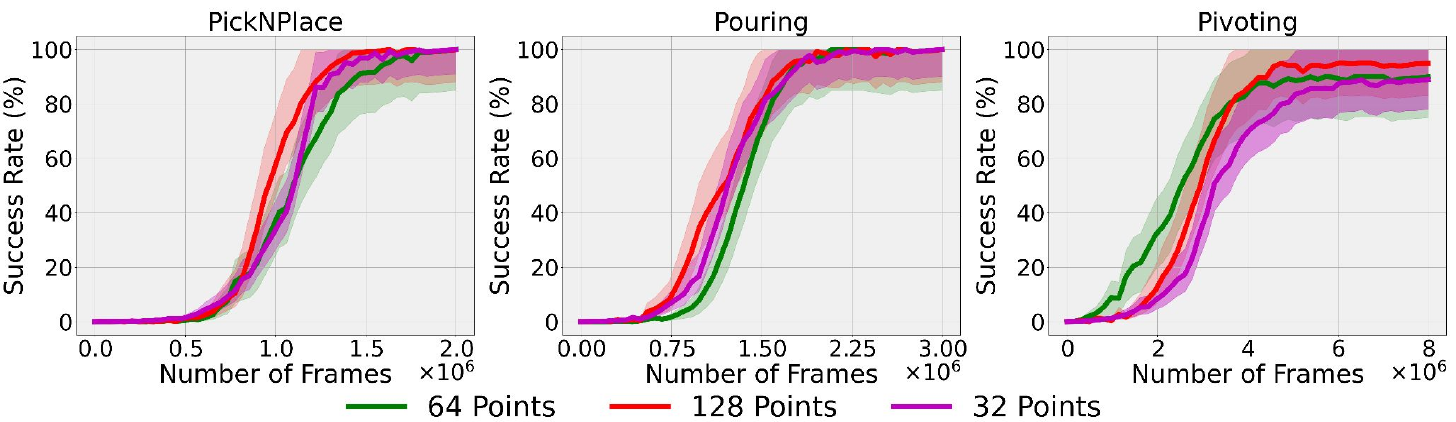}
\vspace{-0.3cm}
    \caption{Results of RL performance with different number of keypoints.}
\label{fig:points}
\vspace{-0.5cm}
\end{figure}

To conduct more evaluations with different number of keypoints, we evaluate the performance with 32 keypoints and 64 keypoints instead of 128 keypoints in the experiments. We find that the performance is similar to using 128 keypoints, which highlights that the stability of the delta-flow fomulations. The results is shown in Fig.~\ref{fig:points}.

\subsection{Ablation Study of Tracking Model}
Compared with Im2Flow2Act~\cite{flow2act}, we propose to use CoTracker~\cite{cotracker} instead of ~\cite{tapir} since Cotracker is better for occupation, which is important for contact-rich manipulation tasks.

To evaluate their performance, we visualize the qualitative results of them for the contact-rich manipulation task \textit{Pivoting}, which is shown in Fig.~\ref{fig:tracker}. Through qualitative results, the CoTracker has much better performance than Tapir when the object meets obstruction.

\begin{figure}[ht]
\vspace{-0.2cm}
    \centering
    \includegraphics[width=0.5\textwidth]{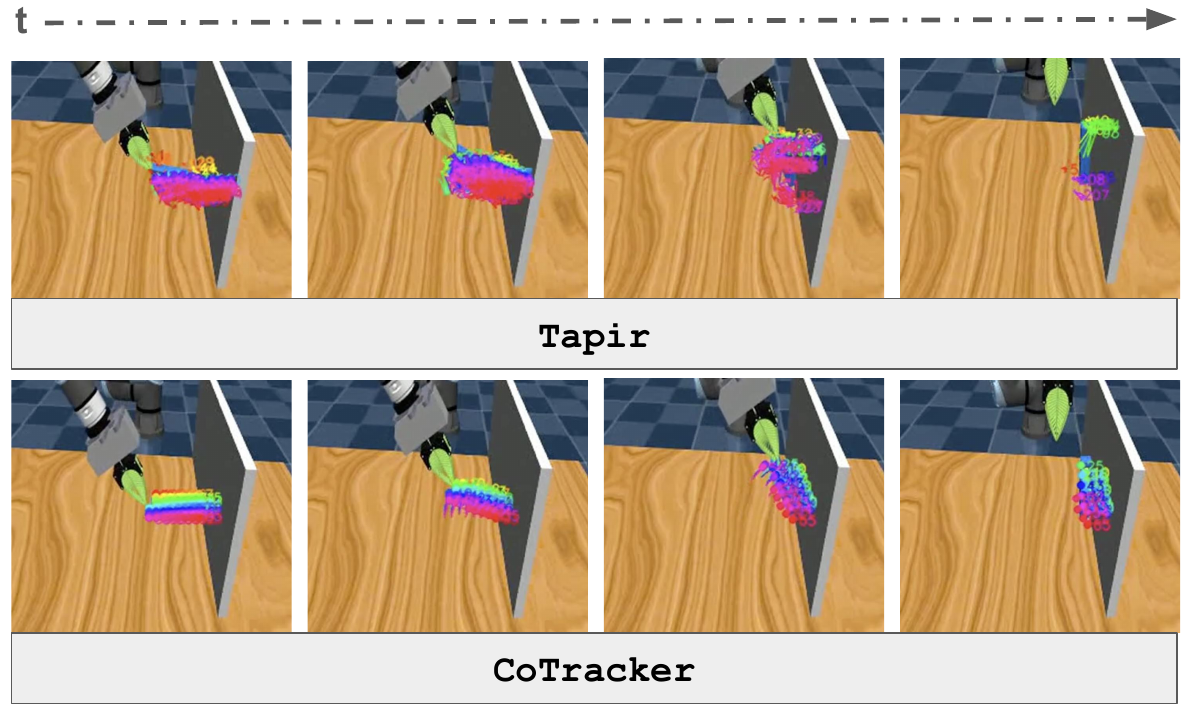}
    \vspace{-0.4cm}
    \caption{Visualization of the comparison of trackers.}
\label{fig:tracker}
\vspace{-0.4cm}
\end{figure}

\section{Real World Case Study Implementation Details}
In this section, we set collect actions from robot actions script with a limit for 200 steps for \textit{Folding} and \textit{PickNPlace}, and 100 steps for \textit{Pouring} and \textit{Pivoting} respectively. The robot control frequency and the CoTracker frequency are both 2.5 Hz. For collecting human demonstrations, the frequency of the CoTracker is 5 Hz, and the final number of steps are the same as the robot demo. Then, we calculate the flow matching reward between the human trajectory and robot trajectory.

\begin{figure*}[t]
\vspace{-0.2cm}
    \centering
    \includegraphics[width=1\textwidth]{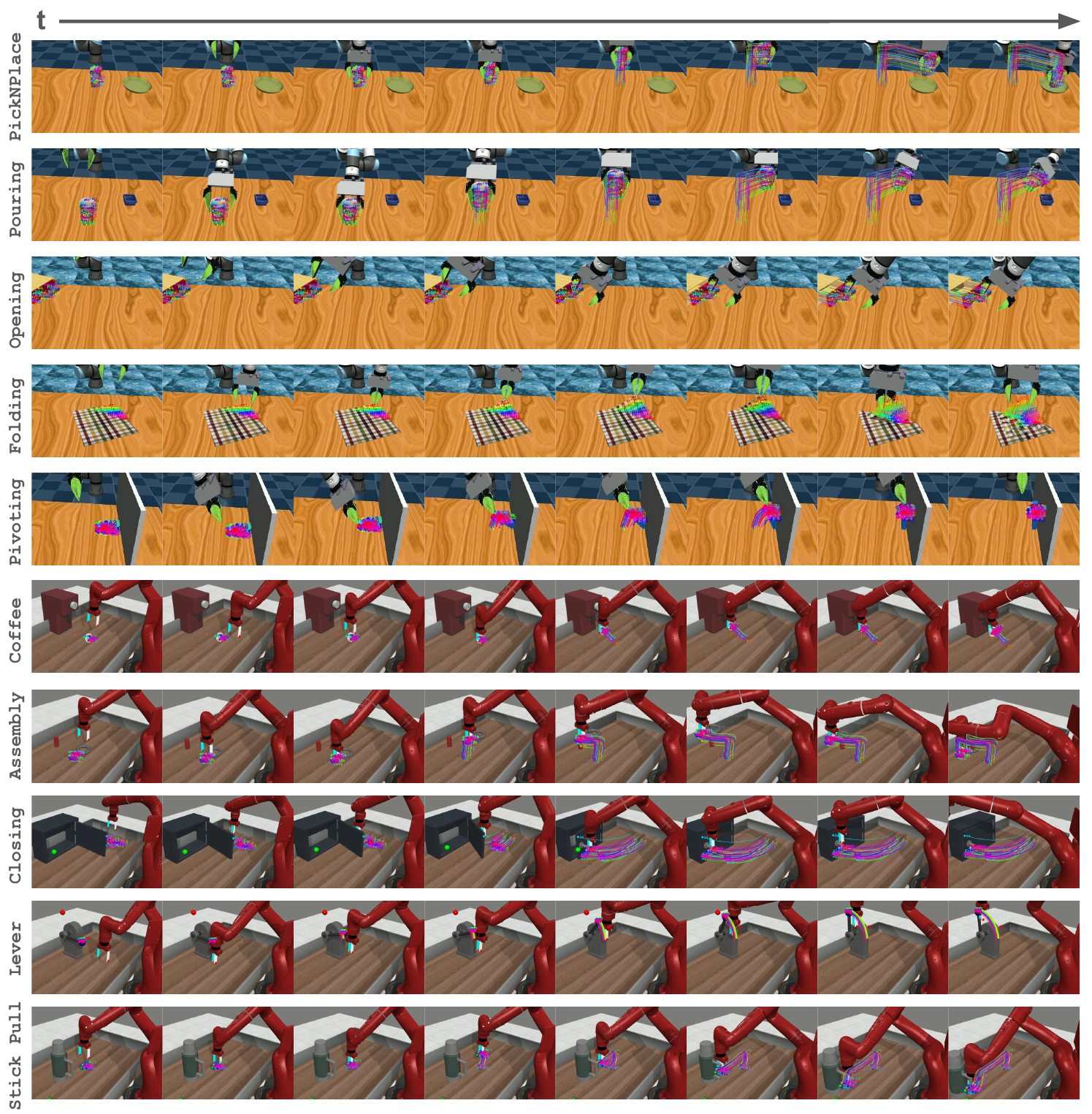}
    \vspace{-0.4cm}
    \caption{The qualitative result of the policy rollout in simulatior.}
\label{fig:tracker}
\vspace{-0.4cm}
\end{figure*}

\begin{figure*}[t]
\vspace{-0.2cm}
    \centering
    \includegraphics[width=1\textwidth]{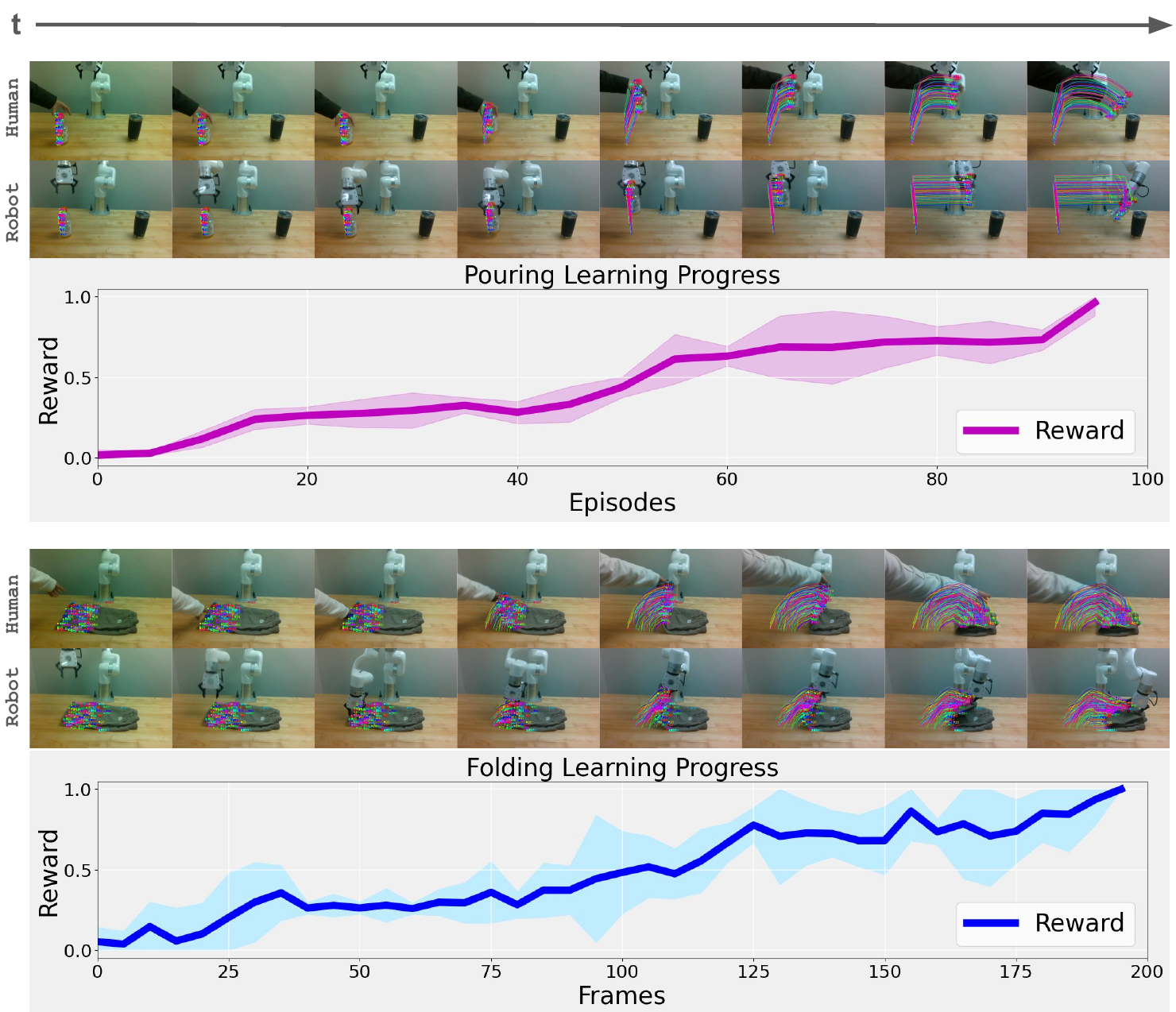}
    \vspace{-0.4cm}
    \caption{The qualitative result of the Flow Matching Reward Case Study in the real world.}
\label{fig:tracker}
\vspace{-0.4cm}
\end{figure*}


\end{document}